\documentclass[5p,times]{elsarticle}

\usepackage{hyperref}

\journal{IFAC Mechatronics journal}

\usepackage{amsmath} 
\usepackage{amssymb}  
\usepackage{url}
\usepackage{graphicx}
\usepackage[font={small}]{caption}
\usepackage[font={small}]{subcaption}
\usepackage{dblfloatfix}
\usepackage{algorithm}
\usepackage{algorithmic}
\usepackage{multirow}
\usepackage{color}
\usepackage{pdfpages}
\usepackage{flushend}
\usepackage{xfrac} 
\usepackage{nicefrac}
\usepackage{wrapfig}

\begin{document}

\begin{frontmatter}

\title{Learning Cooperative Dynamic Manipulation Skills from Human Demonstration Videos}


\author[mymainaddress,mysecondaryaddress]{Francesco Iodice}
\cortext[mycorrespondingauthor]{Corresponding authors}
\ead{francesco.iodice@iit.it }
\author[mymainaddress,mythirdaddress,myfourthaddress]{Yuqiang Wu}
\ead{wuyuqiang@stu.xjtu.edu.cn}

\author[mymainaddress,myfifthaddress]{Wansoo Kim }
\author[mythirdaddress,myfourthaddress]{Fei Zhao}
\author[mysecondaryaddress]{Elena De Momi}
\author[mymainaddress]{Arash Ajoudani}



\address[mymainaddress]{HRI$^{2}$ Lab of Italian Institute of Technology (IIT), Genoa, Italy }
\address[mysecondaryaddress]{Department of Electronics, Information and Bioengineering, Politecnico di Milano, Milano, Italy}
\address[mythirdaddress]{State Key Laboratory for Manufacturing System Engineering, Xi'an Jiaotong University, Xi'an Shaanxi, China }
\address[myfourthaddress]{State Key Laboratory of Intelligent Robots and School of Mechanical Engineering, Xi'an Jiaotong University, Xi'an Shaanxi, China}
\address[myfifthaddress]{Robotics Department, Hanyang University ERICA, Ansan, South Korea.}

\begin{abstract}
This article proposes a method for learning and robotic replication of dynamic collaborative  tasks from  offline videos. The objective is to extend the concept of learning from demonstration (LfD) to dynamic scenarios, benefiting from widely available or easily producible offline videos. To achieve this goal, we decode important dynamic information, such as the Configuration Dependent Stiffness (CDS), which reveals the contribution of arm pose to the arm endpoint stiffness, from a three-dimensional human skeleton model. Next, through encoding of the CDS via Gaussian Mixture Model (GMM) and decoding via Gaussian Mixture Regression (GMR), the robot's Cartesian impedance profile is estimated and replicated. We demonstrate the proposed method in a collaborative sawing task with leader-follower structure, considering environmental constraints and dynamic uncertainties. The experimental setup includes two Panda robots, which replicate the leader-follower roles and the impedance profiles extracted from a two-persons sawing video.
\end{abstract}


\end{frontmatter}


\section{Introduction}

To make robots easily programmable and usable by non-experts, in the last decade, Learning from Demonstration (LfD) has become a major topic in robotics research \cite{argall2009survey,yamaguchi2014learning,fitzgerald2015similarity}. The benefits of LfD over other robot programming methods are more evident when ideal behaviors can neither be scripted (e.g., simple point-to-point trajectories), nor be formulated (e.g., through an optimisation problem or reinforcement learning), but instead can be demonstrated. In such a way, naive workers can easily access and operate robotic systems, and respond to the flexibility requirements of today's custom-made manufacturing \cite{ajoudani2020smart}.

Recently, the advances in LfD with deep neural networks have enabled the learning of complex robot skills involving large and different datasets of raw images. These prominent works in Visual Imitation have shown promising results and demonstrated utility in applications such as manipulating deformable planar objects \cite{zeng2020transporter}, performing intricate manipulations such as pushing, grasping, and stacking \cite{zhang2018deep,zhu2018reinforcement}, and simulating physical forces from videos of humans interacting with objects \cite{ehsani2020use}.  However, although these approaches are promising for modern industrial challenges, they require a huge number of demonstrations, and have limitations in implicitly learning task constraints from demonstrations as they rely solely on extrapolation of skeletal kinematics.

\begin{figure}[!t]
    \centering
    \includegraphics[trim=1.65cm 4.0cm 1.5cm 3.90cm,clip,width=\linewidth]{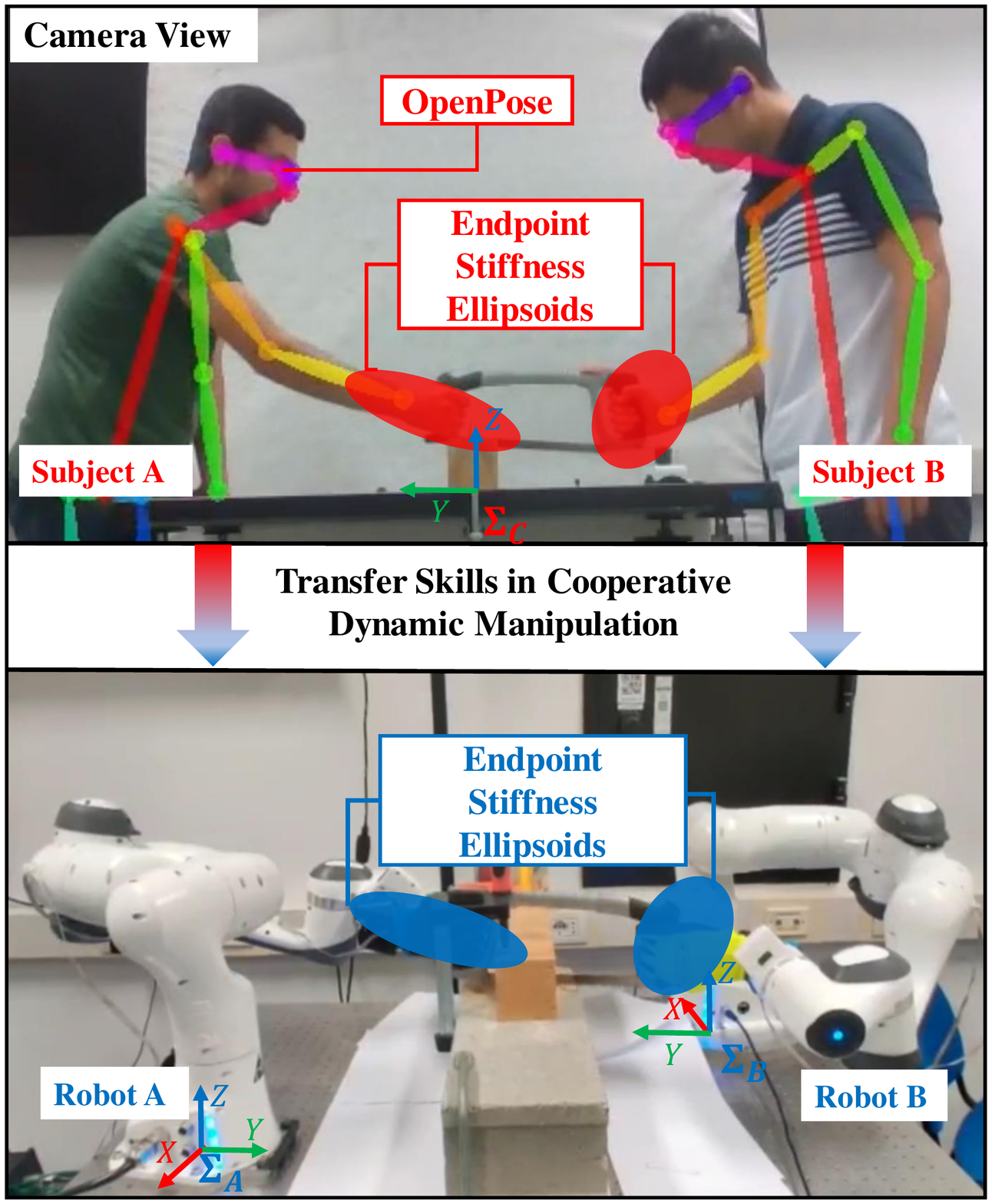}
    \caption{Illustration of the concept and the experimental setup. The objective is to extract leader-follower roles and dynamic features such as configuration-dependent stiffness (CDS) from videos of humans performing collaborative tasks (here, a two-person wood sawing), and replicating them in a dual-arm robotic setup. $\Sigma_C$ represents the camera local frame while $\Sigma_A$ and $\Sigma_B$ represent the base frames of \textit{Robot A} and \textit{B} respectively. The axes of $\Sigma_C$ align with $\Sigma_A$ and $\Sigma_B$.}
    \label{fig:experiment_setup}
    \vspace{-0.6cm}
\end{figure} 

With our work we take into account these challenges and propose a solution for learning collaborative-dynamic manipulation skills from human demonstration videos. Our motivation is to exploit easily producible, large, and diverse video datasets of human operations in industrial environments to decode dynamic features in addition to the leader-follower relations in collaborative tasks. To achieve this, we use an innovative approach in LfD for collecting raw data based on the 3D transformation of the human pose tracking library, OpenPose \cite{cao2018openpose}, after which we rely on the contribution of arm endpoint stiffness profiles, extracted from human demonstrations, to the task dynamics \cite{ajoudani2017choosing}, and their geometric variations, to the leader-follower relations. In this direction, we exploit the Configuration Dependent Stiffness (CDS) model, previously developed in our work \cite{wu2020intuitive}, which represents the dominant contribution of arm configuration to the arm endpoint stiffness geometry \cite{milner2002contribution}. Subsequently, the Gaussian Mixture Model (GMM) and Gaussian Mixture Regression (GMR) are used to encode and reproduce the CDS. Finally, the desired trajectory and stiffness profiles for both robots are achieved by a Cartesian variable impedance controller \cite{wu2020framework}, and the cooperative dynamic two-person sawing skill is transferred from human demonstrated videos to robots.

As a proof-of-concept for collaborative operations, we chose a two-person sawing task that requires effective regulation of physical interaction parameters at hand, in addition to the leader-follower relations \cite{peternel2014teaching}, as depicted in Fig.~\ref{fig:experiment_setup}. This dynamic manipulation task's challenges are proper motion and compliance coordination (leader/follower role allocation) of the two subjects, rejecting external disturbances, and maintaining contact stability while performing the task. 

\begin{figure*}[t!]
    \centering
    \includegraphics[trim=8.0cm 2.0cm 8.0cm 3.0cm,clip,width=\linewidth]{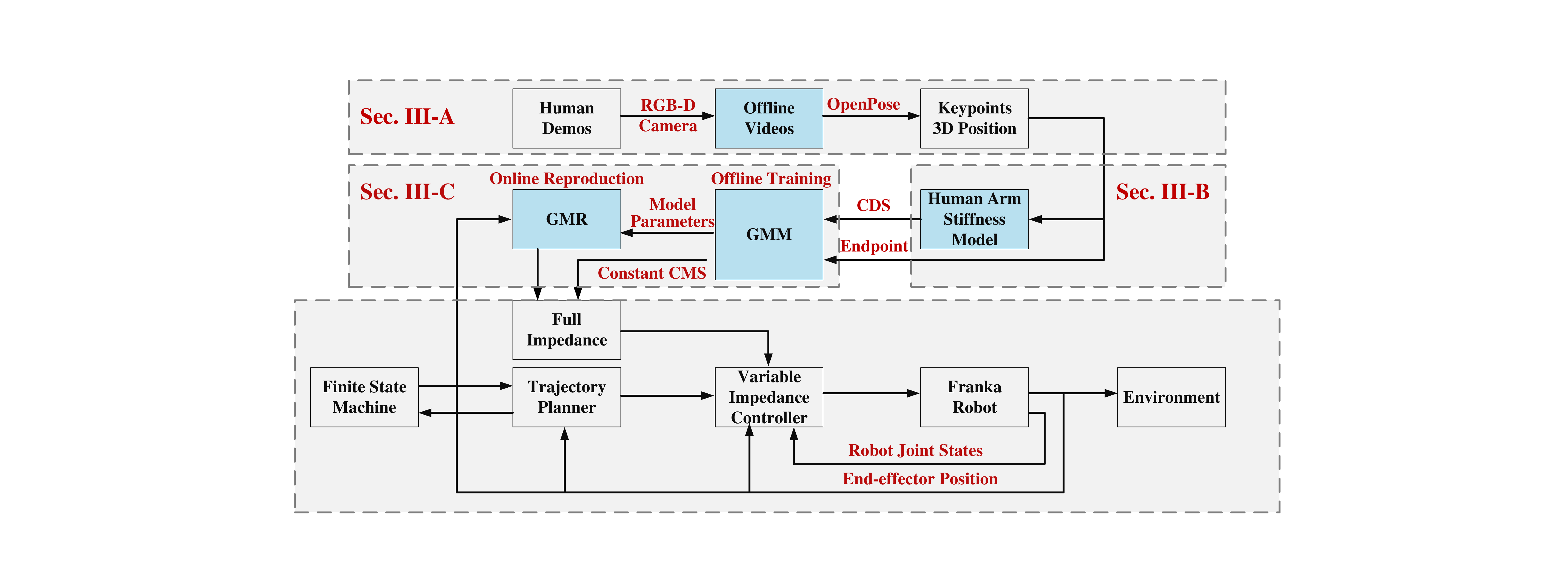}
    \caption{The proposed framework for robots learning dynamic manipulation skills from human demonstration videos. Firstly, human demonstrations are directly recorded by an RGB-D camera and saved as an offline video. Then by using OpenPose, the 3D position information of the keypoints distributed in the human body skeleton model is extracted. Furthermore, the human arm endpoint stiffness model proposed in \cite{wu2020intuitive} is employed to construct the CDS part based on the data extracted from the video. Finally, human demonstrated data is encoded by GMM through offline training and GMR is used to reproduce and generalize CDS information online. The full stiffness matrix, formulated by position-dependent CDS and constant CMS is implemented on robots by a Cartesian impedance controller, whose desired pose is provided by a fifth-order trajectory planner and a Finite State Machine (FSM).}
    \label{fig:framework}
    \vspace{-0.5cm}
\end{figure*}

\section{Related Work}
\label{sec:Related Work}
Most of the Learning from Demonstration methods operate based on one of the following data collection methods: i) kinesthetic teaching, ii) joystick teleoperation, and iii) wearable-based teleoperation. In kinesthetic teaching, the human teacher comes into direct contact with the robot, moving it slowly and reproducing the desired trajectory sequences while the robot records them \cite{calinon2007teacher,kronander2013learning}. This method, which is widely used in motion trajectory learning, helps avoid the correspondence problem caused by the human-robot mapping function. An alternative to kinesthetic teaching is teleoperation, which provides the most direct method to transfer information in demonstration learning. 
During teleoperation, the follower robot is operated by the human leader and the sensory data is recorded. Joystick-based interfaces are very common in such leader-follower settings, whose applications can be found in a variety of cases, including assembly tasks \cite{chen2003programing}, piloting an aerial robot \cite{abbeel2007application,vigouroux2013identification}, obstacle avoidance and navigation \cite{inoue1999acquisition, smart2002making}. However, a different 
type is wearable-based teleoperation, where the robot mimics the motions demonstrated by the teacher while recording data from body sensors. An example of this interface is used by \cite{kohlmorgen2001line} to control the robot pose using the Xsens MVN motion capture inertial suit worn by the user, and by \cite{peternel2016towards}, which is based on an optical tracker and EMG Sensors to develop a multimodal interface to provide the robot with feedback on human motor behavior in real-time. 


Although all three approaches are widely used in LfD applications, when it comes to high degree-of-freedom (DoF) robotic systems or multi-robot cooperation, kinaesthetic teaching and direct interaction interfaces can be time-consuming (due to the need to teach each robot separately).
Similarly, teleoperation interfaces can be associated with high costs and discomfort for humans, as these interfaces may require expensive haptic devices worn by the operators \cite{wu2019teleoperation}.

Due to these limitations and considering practical industrial applications,  the use of cameras to record human movements is encouraged, which brings several advantages: cameras are usually low cost, convenient to deploy, and can be used to create large and diverse datasets for industrial operations. In this respect, deep learning techniques have shown promising results for human modeling after the phase of extrapolating raw data from human performances. The tools used often involve both Reinforcement Learning (RL) and Imitation Learning (IL). For example, using RL-based models, in \cite{peng2018sfv}, the authors were able to train a video prediction model to imitate human poses, and in \cite{sermanet2018time} to imitate human interactions with objects. For IL methods, end-to-end approaches based on mapping from pixels captured in the video to actions have been reported for object manipulation \cite{zeng2020transporter} and the simulation of physical forces in \cite{ehsani2020use}. Some approaches combine RL and IL \cite{balaguer2011combining}; however, these works generally focus on optimizing a specific problem using a task-specific cost function. 



The main limitation of the current vision-based skill imitation techniques is that, learning task dynamics is much less straightforward in comparison to the task kinematics. This limitation becomes even more evident when the plan is to imitate a human demonstrator during manipulation tasks, in which task dynamics contribute majorly to its accomplishment. Human beings can adapt their arm endpoint stiffness ellipsoid's geometry and volume to match the interaction requirements of a dynamic manipulation task \cite{milner2002contribution}. Thus, human arm endpoint stiffness is a preferred representation of task-related interacting dynamics. The online human stiffness estimation techniques were first proposed under the concept of teleimpedance control \cite{ajoudani2012tele}, for real-time transferring of human arm stiffness to teleoperated robots. In \cite{ajoudani2012teleimpedance}, the theory of the Common Mode Stiffness (CMS) and Configuration Dependent Stiffness (CDS) was proposed to explore their roles in realizing desired task space impedance for robot control. Then the idea is used in \cite{ajoudani2015reduced} to map human arm joint stiffness to endpoint stiffness through arm Jacobian, in which CMS reflects the effect of muscular co-activation
and CDS reflects the geometric contribution of arm Jacobian. However, the proposed model can only work within a certain volume of arm workspace due to the configuration-dependent nature of the joint stiffness. To extend this model to a larger arm workspace, the mapping from muscle stiffness to endpoint stiffness is proposed in \cite{ajoudani2018reduced}, Although it resulted in an online and accurate model, however, calculating the muscle Jacobian can be quite complex, hindering its wide spread use in practice. In our recent work, an intuitive human arm endpoint model is proposed based on construction of eigenvectors and eigenvalues of stiffness matrix \cite{wu2020intuitive} which only requires position information of shoulder, elbow and wrist to calculate CDS. It results in a model which is easier to develop and with less parameters. 

The main contribution of this paper to develop a method to learn task dynamics through human demonstration videos. To this end, after extrapolating the raw data from the 3D position estimation obtained by a transformation in OpenPose\cite{cao2018openpose}, we implement a policy, based on a low-level representation approach that takes the form of a non-linear mapping between sensory and motor information, obtaining a modeling action through a probabilistic Gaussian Mixture Model - Gaussian Mixture Regression (GMM-GMR). 


\section{Skills Learning Framework}
\label{sec:Skills Learning Framework}
The overall learning framework is depicted in Fig.~\ref{fig:framework}. 
Firstly, human demonstrations are directly recorded by an RGB-D camera and saved as an offline video. Then by using OpenPose, the 3D position information of the keypoints distributed in the human body skeleton model is extracted. Furthermore, the human arm endpoint stiffness model proposed in \cite{wu2020intuitive} 
is employed to construct the CDS part based on the data extracted from the video. 
Finally, human demonstrated data is encoded by GMM through offline training and GMR is used to reproduce and generalize CDS information online. A constant CMS component is adopted in our framework to incorporate minimum prior knowledge of the task. The full stiffness matrix, formulated by position-dependent CDS and constant CMS is implemented on robots by a Cartesian impedance controller, whose desired pose is provided by a fifth-order trajectory planner and a Finite State Machine (FSM). In the following sections, each part of the overall framework will be explained in more detail.

\subsection{Keypoints information extraction from offline videos}\label{AA}
\label{}
\begin{figure}[t!]
    \centering
    \includegraphics[width=\linewidth]{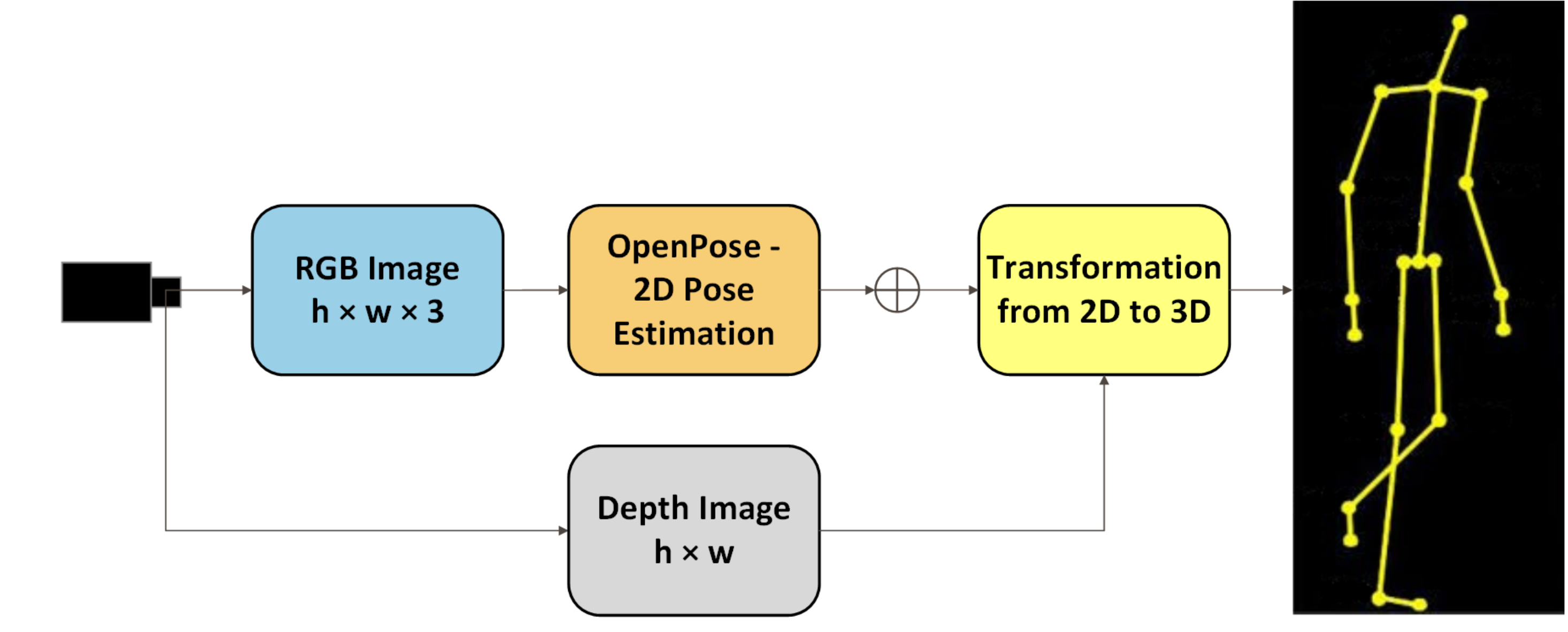}
    \caption{Schematic of the 3D multi-person joint position reconstruction.}
    \label{fig:Openpose Scheme}
    \vspace{-0.5cm}
\end{figure}

In this section, we describe in detail the required stages to transform the 2D human pose data in the 3D space, as depicted in Fig.~\ref{fig:Openpose Scheme}. These data were extracted from a previously recorded learning video, and stored in a rosbag.

\subsubsection{2D Human Pose Estimation}
Human pose estimation is not only defined as the problem of localization of human joints in images or videos, but also as the search for a specific pose in the space of all articulated poses. Over the years, different approaches to human pose estimation have been introduced. The earliest methods were typically estimating the pose of a single person in a static image based on a representation of deformable parts \cite{yang2012articulated}. Of course, these methods are not particularly useful in many real-life scenarios where images contain multi-person. 

In this regard, there are several approaches to estimating the pose of multi-people\cite{toshev2014deeppose,go2016flexible}, which is more complex than the single-person case, as the position and number of people in an image are unknown. In our approach, we use Openpose\cite{cao2018openpose}, one of the most popular bottom-up approaches, that takes, as input, a color image of size w $\times$ h and produces, as output, the 2D locations of anatomical keypoints for each person in the image. It uses a feed-forward network to simultaneously provide both a series of 2D confidence maps of body parts positions and a series of 2D vector fields of the affinities of the parts, which code the degree of association between the parts and then be parsed by greedy inference to output the 2D keypoints for all people in the image.
The extracted information is subsequently used to create a configuration of the skeleton model, with the different number of body joints changing according to the dataset used (in our case, we have 25 keypoints consisting of 18 body keypoints by COCO dataset \cite{lin2014microsoft} and 6 foot keypoints \cite{cao2018openpose} ). This approach provides reliability in identifying errors for proximity and occlusion, ensuring accurate position estimation.\\

\subsubsection{3D Human Pose Estimation}
To formulate CDS information, we concentrate on reconstructing the 3D pose and implementing a 3D extractor from the 2D positions of the human joints in the RGB image emitted by the OpenPose network, by taking into account the depth information. When the depth image is synchronized with the RGB image, we obtain an image RGB-D, where one can directly extract the corresponding pose estimation in the 3D Cartesian space for each 2D pixel (u,v).
Since the conversion, we pay attention to simultaneously accessing the color and depth values for each pixel of the input RGB-D data, keeping the data organized, setting both resolutions to 640 $\times$ 480 and frequency of 20Hz, to have a one-to-one correspondence depth and color data.
We address occlusion problems during 3D mapping removing all those pixels without depth values (NaN) in the associated point cloud. Then, assuming that the pixels in the 2D human joint position are mapped to the correct depth pixels, we use the closest proximity pixels to average their values.

\subsection{Human arm endpoint stiffness model}
\label{subsec:CDS}
\label{sec:human arm stiffness model}
\begin{figure}[b!]
    \vspace{-0.50cm}
    \centering
    \includegraphics[trim=1.5cm 5.5cm 1.5cm 4.0cm,clip,width=0.90\linewidth]{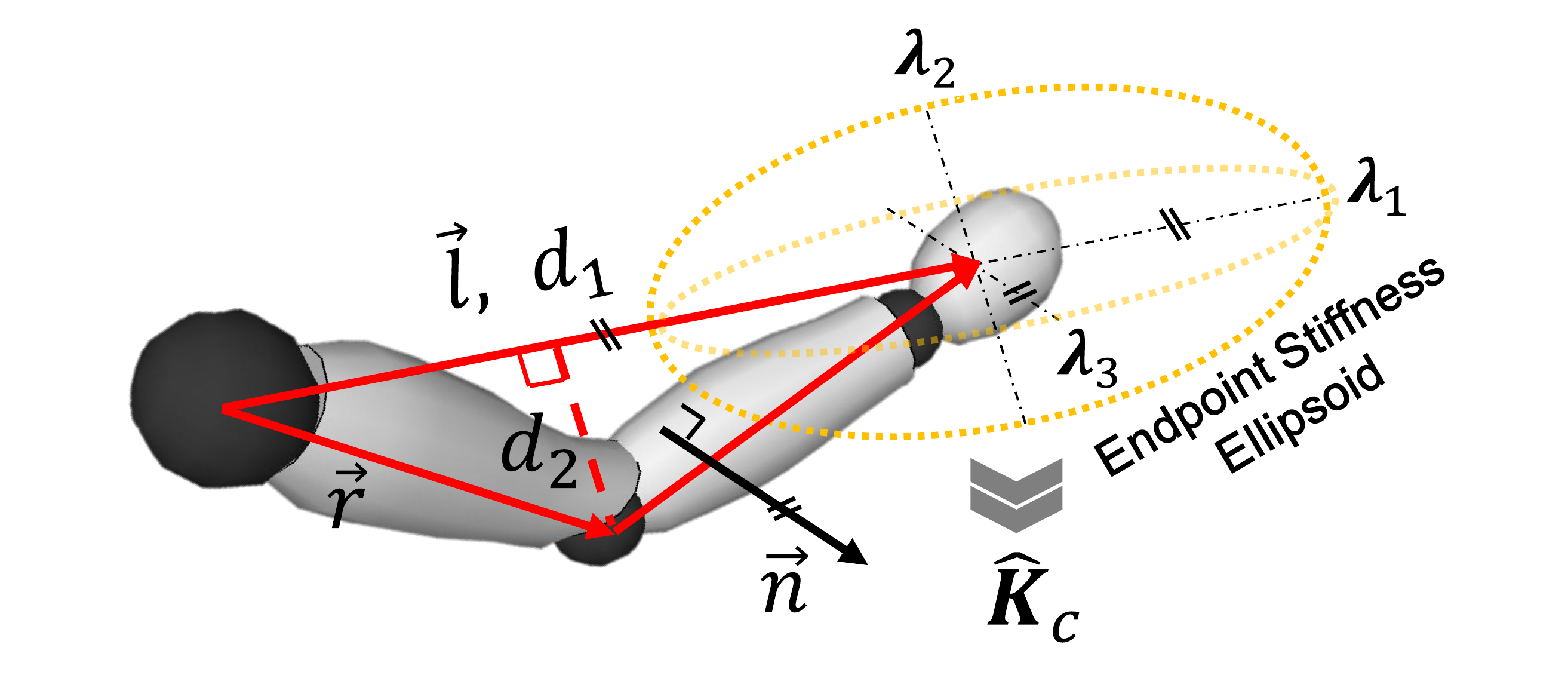}
    \caption{The geometric relationship between the human arm configuration and the principal axes of the endpoint stiffness ellipsoid. $\Vec{\boldsymbol{n}}$ is perpendicular to the plane formed by $\Vec{\boldsymbol{l}}$ and $\Vec{\boldsymbol{r}}$. $\Vec{\boldsymbol{l}}$ and $\Vec{\boldsymbol{n}}$ are parallel to the major and minor principal axes of the endpoint stiffness ellipsoid, respectively.}
    \label{fig: human arm geometry}
    \vspace{-0.00cm}
\end{figure}
The idea of CMS and CDS is inspired by human motor control principles on the predominant use of the arm configuration in directional adjustments of the endpoint stiffness profile, and the synergistic effect of muscular activations, which contributes to a coordinated modification of the endpoint stiffness in all Cartesian directions. In \cite{wu2020intuitive}, we propose a principally simplified, and intuitive online model of the human arm endpoint stiffness. The new model is based on the large dependency of the shape and orientation of the stiffness ellipsoid on the arm configuration. In fact, previous studies \cite{mussa1985neural,zatsiorsky2002kinetics} indicate that i) the major principal axis of the arm endpoint stiffness ellipse passes close to the hand-shoulder vector, and ii) when the arm is extended and the hand moves further from the shoulder, the ellipse becomes more elongated and conversely, it becomes more isotropic. Inspired by these findings, the new model in 3D is constructed by the ellipsoid's principal axes and their lengths, namely the eigenvectors and eigenvalues of the stiffness matrix, based on the arm configuration. 

As shown in Fig.~\ref{fig: human arm geometry}, we use a two-segment human arm skeleton structure in the 3D space. The hand-forearm and the upper arm segments in this model permit us to form an arm triangle at any non-singular configuration. Relying on the dominant contribution of the arm configuration to the endpoint stiffness geometry, we propose to use the vector from the centre of shoulder joint to the position of the hand ($\Vec{\boldsymbol{l}}\in\mathbb{R}^{3}$), to identify the major principal direction of the human arm endpoint stiffness ellipsoid. The minor principal axis direction ($\Vec{\boldsymbol{n}}\in\mathbb{R}^{3}$), instead, is defined to be perpendicular to the arm triangle plane
\begin{equation}
    \Vec{\boldsymbol{n}} = \Vec{\boldsymbol{r}} \times \Vec{\boldsymbol{l}},
\end{equation} 
where $\Vec{\boldsymbol{r}}\in\mathbb{R}^{3}$ represents the vector from the centre of shoulder to the centre of elbow. The remaining principal axis of the stiffness ellipsoid, which lies on the arm triangle plane, is calculated based on the orthogonality of the three principal axes. Under the assumption that the ratio of the length of median principal axis to the major principal axis of the stiffness ellipsoid is inversely proportional to the distance $d_1 \in\mathbb{R}$, from the hand position to the centre of shoulder, while the ratio of the length of the minor principal axis to the major principal axis is proportional to the distance $d_2 \in\mathbb{R}$, from the centre of the elbow to the major principal axis, the estimated endpoint stiffness matrix $\boldsymbol{\hat{K}}_c\in\mathbb{R}^{3 \times 3}$ is formulated by
\begin{equation}
\label{eq:model}
    \boldsymbol{\hat{K}}_c =  \boldsymbol{V}A_{cc}\boldsymbol{D}_s\boldsymbol{V}^T.
\end{equation}
Here, $A_{cc}\in\mathbb{R}$ is the co-contraction activation index of human arm muscles corresponding to CMS, while $\boldsymbol{V}\boldsymbol{D}_s\boldsymbol{V}^T \in \mathbf{R}^{3 \times 3}$ corresponding to CDS, and $\boldsymbol{V}\in\mathbb{R}^{3 \times 3}$ and $\boldsymbol{D}_s\in\mathbb{R}^{3 \times 3}$ are respectively eigenvectors representing the orientation and eigenvalues representing the shape of the stiffness ellipsoid and formulated by:
\begin{equation}
    \boldsymbol{V} = \Bigg [\frac{\Vec{\boldsymbol{l}}}{\parallel \Vec{\boldsymbol{l}} \parallel}, \frac{(\Vec{\boldsymbol{r}} \times \Vec{\boldsymbol{l}}) \times \Vec{\boldsymbol{l}}}{\parallel (\Vec{\boldsymbol{r}} \times \Vec{\boldsymbol{l}}) \times \Vec{\boldsymbol{l}} \parallel}, \frac{\Vec{\boldsymbol{r}} \times \Vec{\boldsymbol{l}}}{\parallel \Vec{\boldsymbol{r}} \times \Vec{\boldsymbol{l}} \parallel} \Bigg ],
    \label{eq:eigen_vectors}
\end{equation}
\begin{equation}
    \boldsymbol{D}_s =
    \frac{\text{diag} (1, \quad \alpha_1 / d_1, \quad \alpha_2 d_2)}{ (1 \times \alpha_1 / d_1 \times \alpha_2 d_2)^{\frac{1}{3}}},
\end{equation}
where $\alpha_1 \in\mathbb{R}$ and $\alpha_2 \in\mathbb{R}$ are subject related parameters in the model. For the detailed explanation and further model parameter identification, please refer to \cite{wu2020intuitive}. 

In most applications, the stiffness geometry (CDS) plays a dominant role compared to its volume (CMS) \cite{wu2020framework}. Considering our learning framework is only based on vision sensors while CMS can only be recorded by electromyography (EMG) sensors, a suitable constant CMS component will be selected according to the allowable range of translational stiffness of robots to involve minimum prior knowledge of the tasks. According to the introduced model, to extract CDS information from human demonstrations for a given task, only positional information of the shoulder, elbow and wrist is necessary which could be collected even from an offline video. The CDS information can be extracted followed by a post-processing of the collected 3D position data according to the proposed model. 

\subsection{Skills transfer by GMM/GMR}
\label{subsec:Skills Transfer by Imitation Learning}
In this part, we first give a recap of GMM/GMR
algorithm and then introduce how to apply it to encode the CDS profiles.
The Gaussian mixture distribution can be written as a linear superposition of Gaussians in the form
\begin{equation}
p(\boldsymbol{\xi}) = \sum_{k=1}^{K}\pi_k\mathcal{N}(\boldsymbol{\xi}|\boldsymbol{\mu}_k,\boldsymbol{\Sigma}_k),
\end{equation}
where $\boldsymbol{\xi} \in\mathbb{R}^{d}$ and $p(\boldsymbol{\xi}) \in\mathbb{R}$, respectively, represent the vector of variables and joint probability distribution. 
$\pi_k \in\mathbb{R}$, $\boldsymbol{\mu}_k\in\mathbb{R}^{d}$ and $\boldsymbol{\Sigma}_k\in\mathbb{R}^{d \times d}$ represent the prior probability, mean and covariance of the $k$-th Gaussian component respectively, while $K\in\mathbb{Z}^+$ is the total number of Gaussian components. 

The goal of the offline training phase is to maximize the log likelihood function Eq.~(\ref{eq:log_likelihood}) with respect to the model parameters ($\pi_k$, $\boldsymbol{\mu}_k$ and $\boldsymbol{\Sigma}_k$), which results in an EM procedure (E-step in (\ref{eq:e_step}) and M-step in (\ref{eq:m_step1}) and (\ref{eq:m_setp2})) to iteratively update the model parameters until convergence
\begin{equation}
    \ln p([\boldsymbol{\xi}_1, \boldsymbol{\xi}_2, \boldsymbol{\dots}, \boldsymbol{\xi}_N]) = \sum_{n=1}^N\ln
    \bigg \{\sum_{k=1}^K\pi_k\mathcal{N}(\boldsymbol{\xi}_n|\boldsymbol{\mu}_k,\boldsymbol{\Sigma}_k) \bigg \},
    \label{eq:log_likelihood}
\end{equation}

\begin{equation}
    \gamma_{n,k} = \frac{\pi_k\mathcal{N}(\boldsymbol{\xi}_n|\boldsymbol{\mu}_k,\boldsymbol{\Sigma}_k)}{\sum_{j=1}^K\pi_j\mathcal{N}(\boldsymbol{\xi}_n|\boldsymbol{\mu}_j,\boldsymbol{\Sigma}_j)},
    \label{eq:e_step}
\end{equation}

\begin{equation}
    \pi_k^{new} = \frac{\sum_{n=1}^N\gamma_{n,k}}{N},\quad  
    \boldsymbol{\mu}_k^{new} = \frac{\sum_{n=1}^{N}\gamma_{n,k}\boldsymbol{\xi}_n}{\sum_{n=1}^N\gamma_{n,k}},
    \label{eq:m_step1}
\end{equation}
\begin{equation}
    \boldsymbol{\Sigma}_k^{new} = \frac{\sum_{n=1}^{N}\gamma_{n,k}(\boldsymbol{\xi}_n - \boldsymbol{\mu}_k^{new})(\boldsymbol{\xi}_n - \boldsymbol{\mu}_k^{new})^T}{\sum_{n=1}^N\gamma_{n,k}}.
    \label{eq:m_setp2}
\end{equation}
Here, $N \in \mathbb{Z}^+$ is the total number of data points in a training dataset and $\boldsymbol{\xi}_n \in \mathbb{R}^{d}$ is the $n$-th training data point.

After modeling the joint probability distribution of the training data offline, to derive GMR, we use superscripts $\mathcal{I}$ and $\mathcal{O}$ to denote the dimensions of the input and output variables. At iteration $n$, $\boldsymbol{\eta}_n^{\mathcal{I}}$ and $\boldsymbol{\eta}_n^{\mathcal{O}}$  represent the input and output variables, respectively. With this notation, the data point $\boldsymbol{\eta}_n \in \mathbb{R}^{d}$, mean $\boldsymbol{\mu}_k$ and covariance $\boldsymbol{\Sigma}_k$ of the $k$-th Gaussian component can be decomposed as 
\begin{equation}
    \boldsymbol{\eta}_n =
    \begin{bmatrix}
    \boldsymbol{\eta}_n^{\mathcal{I}} \\
    \boldsymbol{\eta}_n^{\mathcal{O}}
    \end{bmatrix},
    \boldsymbol{\mu}_k = 
    \begin{bmatrix}
    \boldsymbol{\mu}_k^{\mathcal{I}} \\
    \boldsymbol{\mu}_k^{\mathcal{O}}
    \end{bmatrix},
    \boldsymbol{\Sigma}_k = 
    \begin{bmatrix}
    \boldsymbol{\Sigma}_k^{\mathcal{I}\mathcal{I}} & \boldsymbol{\Sigma}_k^{\mathcal{I}\mathcal{O}} \\
    \boldsymbol{\Sigma}_k^{\mathcal{O}\mathcal{I}} & \boldsymbol{\Sigma}_k^{\mathcal{O}\mathcal{O}}
    \end{bmatrix}.
\end{equation}

In the online reproduction phase, the best estimation of output $\boldsymbol{\hat{\eta}}_n^{\mathcal{O}}$ for a given input $\boldsymbol{\eta}_n^{\mathcal{I}}$ is the mean $\boldsymbol{\hat{\mu}}_n$ of the conditional probability distribution $\boldsymbol{\hat{\eta}}_n^\mathcal{O}|\boldsymbol{\eta}_n^\mathcal{I} \sim \mathcal{N}(\boldsymbol{\hat{\mu}}_n, \boldsymbol{\hat{\Sigma}}_n)$, with parameters which was reported in \cite{ghahramani1994supervised}:
\begin{equation}
    \boldsymbol{\hat{\mu}}_n = \mathbb{E}(\boldsymbol{\hat{\eta}}_n^\mathcal{O}|\boldsymbol{\eta}_n^\mathcal{I}) = \sum_{k=1}^{K}h_k(\boldsymbol{\eta}_n^\mathcal{I})\boldsymbol{\mu}_k(\boldsymbol{\eta}_n^\mathcal{I}),
\end{equation}
where 
\begin{equation}
    h_k(\boldsymbol{\eta}_n^\mathcal{I}) = \frac{\pi_k\mathcal{N}(\boldsymbol{\eta}_n^\mathcal{I}|\boldsymbol{\mu}_k^{\mathcal{I}},\boldsymbol{\Sigma}_k^{\mathcal{I}\mathcal{I}})}{\sum_{j=1}^K\pi_j\mathcal{N}(\boldsymbol{\eta}_n^\mathcal{I}|\boldsymbol{\mu}_j^{\mathcal{I}},\boldsymbol{\Sigma}_j^{\mathcal{I}\mathcal{I}})},
\end{equation}
\begin{equation}
    \boldsymbol{\mu}_k(\boldsymbol{\eta}_n^\mathcal{I}) = \boldsymbol{\mu}_k^\mathcal{O}+\boldsymbol{\Sigma}_k^{\mathcal{O}\mathcal{I}}(\boldsymbol{\Sigma}_k^{\mathcal{I}\mathcal{I}})^{-1}(\boldsymbol{\eta}_n^\mathcal{I}-\boldsymbol{\mu}_k^{\mathcal{I}}).
\end{equation}

\begin{figure*}[t!]
    \centering
    \includegraphics[trim=2.0cm 5.75cm 2.0cm 5.50cm,clip,width=\linewidth]{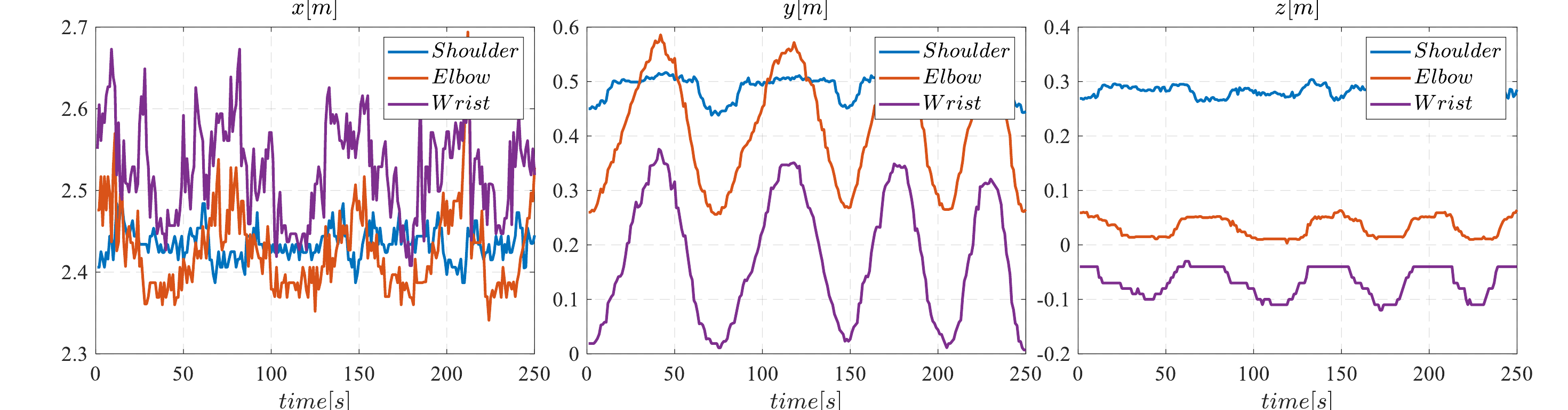}
    \caption{Subplots from left to right respectively correspond to $x$, $y$ and $z$ positions in the camera frame $\Sigma_C$ of keypoints of shoulder (blue), elbow (red) and wrist (purple) of subject A's right arm extracted from the offline video. Results of subject B, which are similar, are not reported to avoid duplication.}
    \label{fig:raw_data_from_video}
    \vspace{-0.5cm}
\end{figure*}

In this framework, only the mean value is used although covariance can be derived as well. Human arm endpoint pose and CDS are respectively considered as input and output of the GMM. Pose data can be represented in vector form, whereas since the CDS profile is a series of Symmetric Positive Definite (SPD) matrices, it's not possible to apply GMM to encode it directly considering GMM can only deal with independent variables in vector form. In \cite{abu2018force}, two different ways to represent stiffness matrix are provided. In the following, Cholesky decomposition is used to decompose CDS into the product of a lower triangular matrix and its transpose, and then the non-zero elements are arranged in the vector form. 
\begin{equation}
	\boldsymbol{K}_{demo} = \boldsymbol{L}\boldsymbol{L}^T,
\end{equation}
where $\boldsymbol{K}_{demo} \in\mathbb{R}^{3 \times 3}$ is the demonstrated stiffness by human and $\boldsymbol{L} \in\mathbb{R}^{3 \times 3}$ is the lower triangular matrix after Cholesky decomposition. The format of $\boldsymbol{L}$ is as following:
\begin{equation}
	\boldsymbol{L} = 
	\begin{bmatrix}
		l_{11} & 0 & 0 \\
		l_{21} & l_{22} & 0 \\
		l_{31} & l_{32} &  l_{33} \\
	\end{bmatrix},
\end{equation}
where $l_{11}$, $l_{22}$ and $l_{33}$ are all positive values. Rearrange the non-zero and independent elements in $\boldsymbol{L}$ into vector form:
\begin{equation}
	\hat{\boldsymbol{L}} =
	\begin{bmatrix}
		l_{11} & l_{21} & l_{22} & l_{31} & l_{32} & l_{33}
	\end{bmatrix} ^T.
\end{equation}
After the above process of demonstrated stiffness matrices,  a series of $\hat{\boldsymbol{L}}$ is obtained, which can be encoded directly by GMM, the offline training phase takes the compact pose and $\hat{\boldsymbol{L}}$  as the vector of variables $\boldsymbol{\xi}$. Subsequently, the GMR reproduces the CDS vector online by taking robot end-effector pose as input. The CDS matrices can be constructed followed by a reverse process of Cholesky decomposition and in this way, SPD property of stiffness matrices is guaranteed. 

\section{Experiments and Results}
\label{sec:Experiments and Results}
As depicted in Fig.~\ref{fig:experiment_setup}, two subjects used a two-person cross cut saw to perform the wood sawing task. The challenges of this dynamic manipulation task are proper motion and impedance coordination (for interaction control and leader/follower role allocation) of the two subjects, to comply with external disturbances and maintain contact stability during the task. In the following, based on the data collected through OpenPose from human demonstration videos and CDS construction method proposed in Sec.~\ref{subsec:CDS}, the motor behaviour of the two subjects in the experiment is first analyzed. Furthermore, the sawing task-related dynamic manipulation skill, extracted from human motions, is transferred to robots through GMM/GMR. The proposed skills learning framework is then validated by comparing to different stiffness parameter settings.

\subsection{Human demonstration learning and analysis}

To learn human arm cooperative dynamic manipulation skills from videos, first, an RGB-D camera was used to record the process and save it as an offline video. After processing the offline video as explained in Sec.~\ref{AA} by using OpenPose, the raw 3D position information of keypoints distributed on the skeleton model of the two subjects can be collected. Here, only the information of keypoints on human right arm was used. The raw shoulder (blue), elbow (red) and wrist (purple) position information from subject A are illustrated in Fig.~\ref{fig:raw_data_from_video}. As can be seen in these plots, the main position variation happened in elbow and wrist keypoints and in $y$ direction in camera local frame $\Sigma_C$, which was the sawing direction.

\begin{figure}
    \centering
    \begin{subfigure}[b]{0.25\textwidth}
        \centering
        \includegraphics[trim=10.75cm 0.0cm 10.75cm 0.0cm,clip,width=\linewidth]{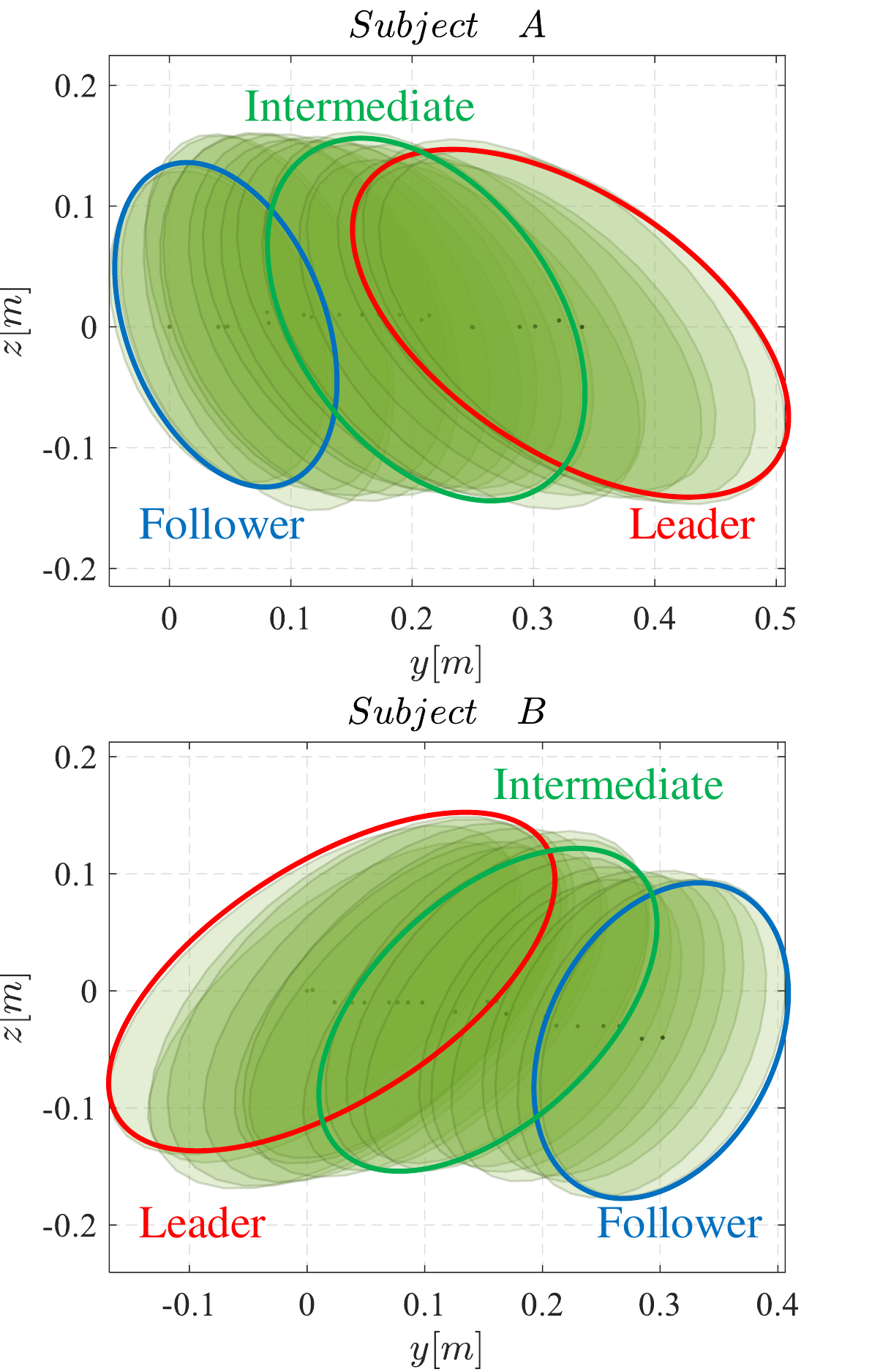}
        \caption{CDS demos for GMM training.}
        \label{fig:cds_demonstrations}
    \end{subfigure}%
    \begin{subfigure}[b]{0.25\textwidth}
        \centering
        \includegraphics[trim=10.75cm 0.0cm 10.75cm 0.0cm,clip,width=\linewidth]{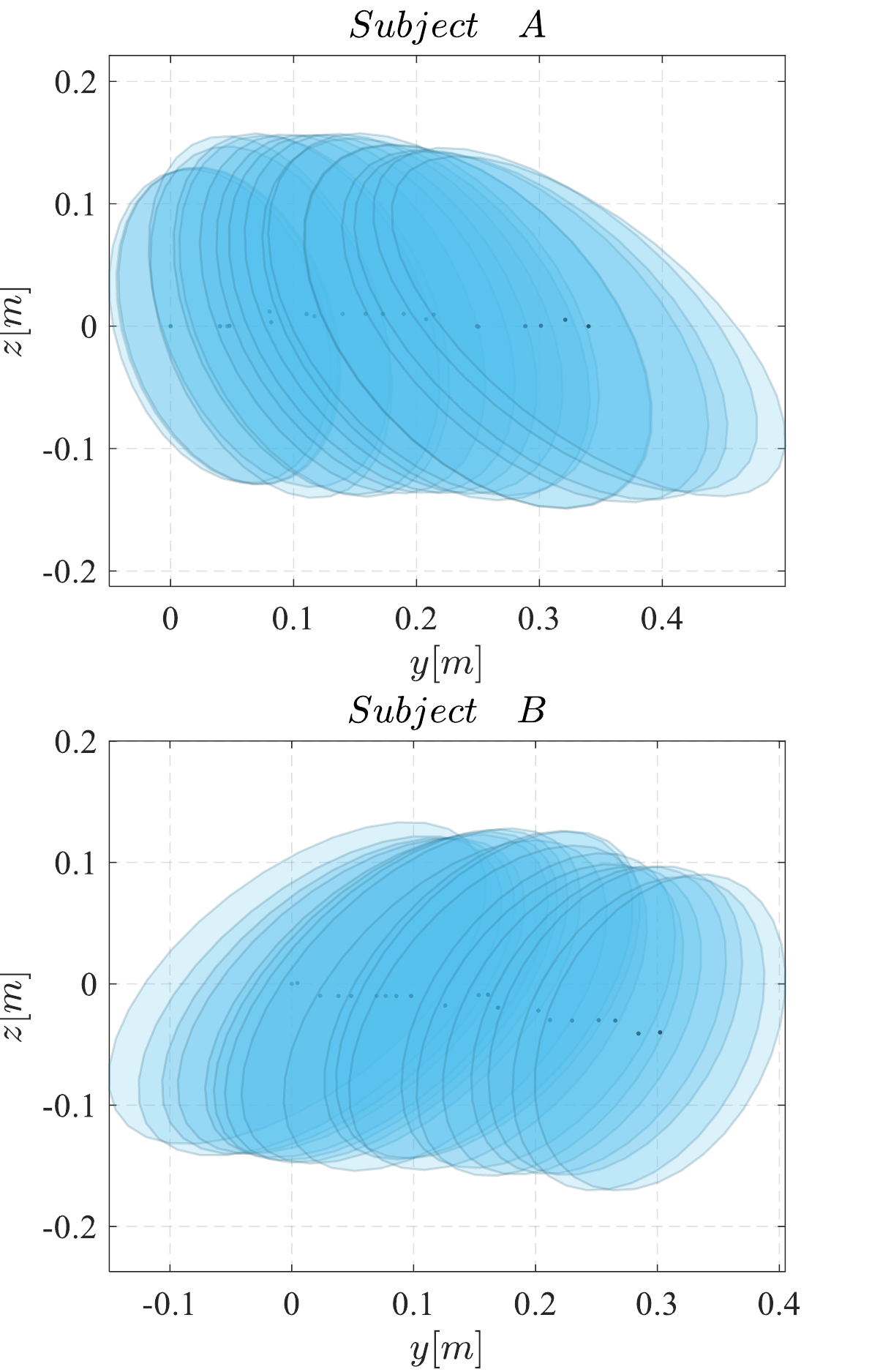}
        \caption{CDS regression from GMR.}
        \label{fig:cds_regressions}
    \end{subfigure}
    \caption{The GMM (green) encoded, and the GMR (blue) decoded CDS profiles of subjects A and B, considering the task plane positions ($y$ and $z$ axis) as the input.}
    \vspace{-0.5cm}
\end{figure}

\begin{figure*}[t!]
    \centering
    \includegraphics[trim=2.0cm 5.5cm 2.0cm 5.5cm,clip,width=\linewidth]{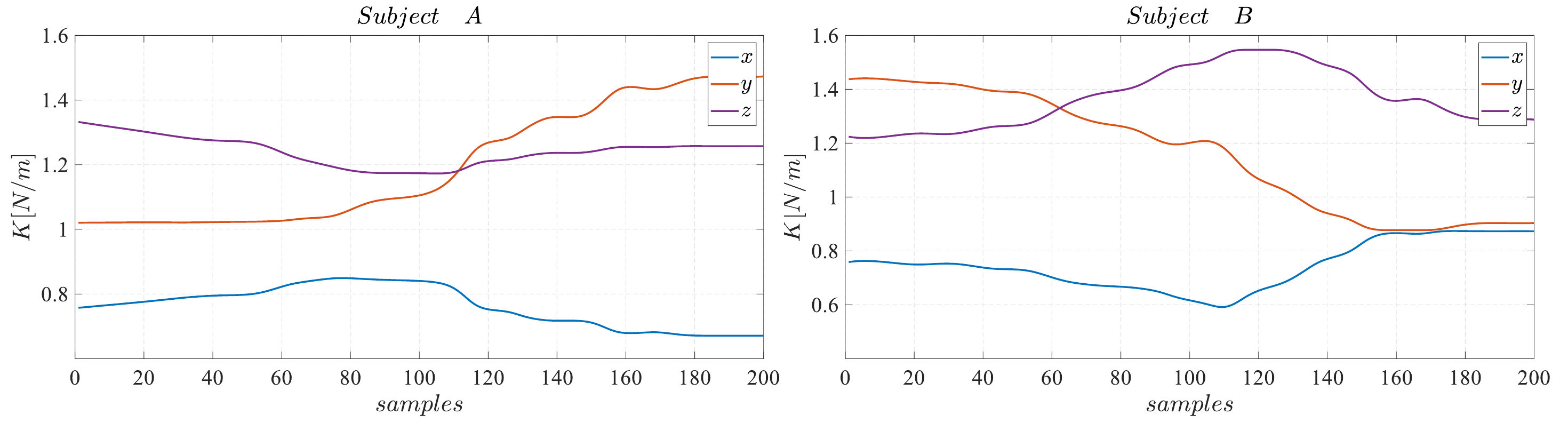}
    \caption{The projected CDS profiles of the two subjects in $x$, $y$ and $z$ axes, in one leading (subject B)-following (subject A) phase.}
    \label{fig:decoupled_xyz}
    \vspace{-0.5cm}
\end{figure*}

According to the raw human arm keypoints position data extracted from the offline video, by adopting the method proposed in Sec.~\ref{subsec:CDS}, the behaviours of CDS of the two subjects during the cooperative sawing task were constructed. These are  depicted in Fig.~\ref{fig:cds_demonstrations}, in which the 3D stiffness ellipsoids are projected onto $y-z$ plane due to the fact that the movement of the sawing task was constrained in $y-z$ plane. More importantly, we consider that the task-related CDS is position-dependent and does not depend on the execution time. To encode the CDS information extracted from human demonstrations, to reproduce and to generalize it online, the widely used imitation learning algorithm, GMM/GMR was used, which was explained in detail in Sec.~\ref{subsec:Skills Transfer by Imitation Learning}.
By employing the processed data in GMM model training (for both subjects, $y$ and $z$ positions of the subject's wrist as input vector, the corresponding CDS matrix as the output), the priors, mean, and covariance for each Gaussian element were converged. To eliminate the effects of different sawing speed and jitter observed in the raw data collected from the offline video as shown in Fig.~\ref{fig:raw_data_from_video}, a resampling procedure was performed to smooth the raw data and align the length of samples in each demonstration. 
Fig.~\ref{fig:cds_regressions} illustrates the online CDS results, which were reproduced via GMR, the input data was chosen from the demonstration.

An important observation is the change of the stiffness ellipsoids' geometries towards the end of each sawing action (i.e., pulling or pushing) for subjects A and B in Fig. \ref{fig:cds_demonstrations}, which reveals the leader-follower relationship of the involved subjects. In more details, the leading role is played by the subject with more elongated stiffness in the direction of sawing (with an extended arm that contributes to a larger CDS axis), since it can generate larger reactive force under the same tracking error due to the impedance control law. On the other hand, the follower has a more compliant profile in that instant, facilitating the leader's role in pulling the saw to perform the cutting. Hence, such a leader-follower relation, which is repeated in each task cycle, can be easily implemented by reproducing the CDS profiles by the robots.


To make a deeper analysis of the CDS variations extracted from human data, the  reproduced CDS profiles of subjects A and B were projected into three orthogonal directions $x$, $y$ and $z$, as plotted in Fig.~\ref{fig:decoupled_xyz}. It can be seen that, compared with the stiffness profile in $y$ direction, the stiffness in $x$ and $z$ directions didn't show obvious variations, coinciding with the fact that the main task direction is along the $y$ (sawing) axis. Furthermore, the stiffness in $x$ direction during the whole process was relatively lower than the stiffness in the other two directions, while the stiffness in $z$ direction was kept at a relatively high level during the whole process. The reasons behind these observations are the following. Considering that the movement in $x$ direction was severely constrained during the sawing action, a compliant behaviour was achieved to avoid the generation of large reactive forces. In $z$ direction instead, the descending movement of the saw was blocked by the wood, however, the ascending movements could cause an instability. A higher stiffness value in $z$ direction can serve to maintain the stability of the saw during such a highly dynamic manipulation task, whose effect is demonstrated in the following experiments on two robots.

\begin{figure}[b!]
    \vspace{-0.7cm}
    \centering
    \includegraphics[trim=8.5cm 0.5cm 8.5cm 0.5cm,clip,width=\linewidth]{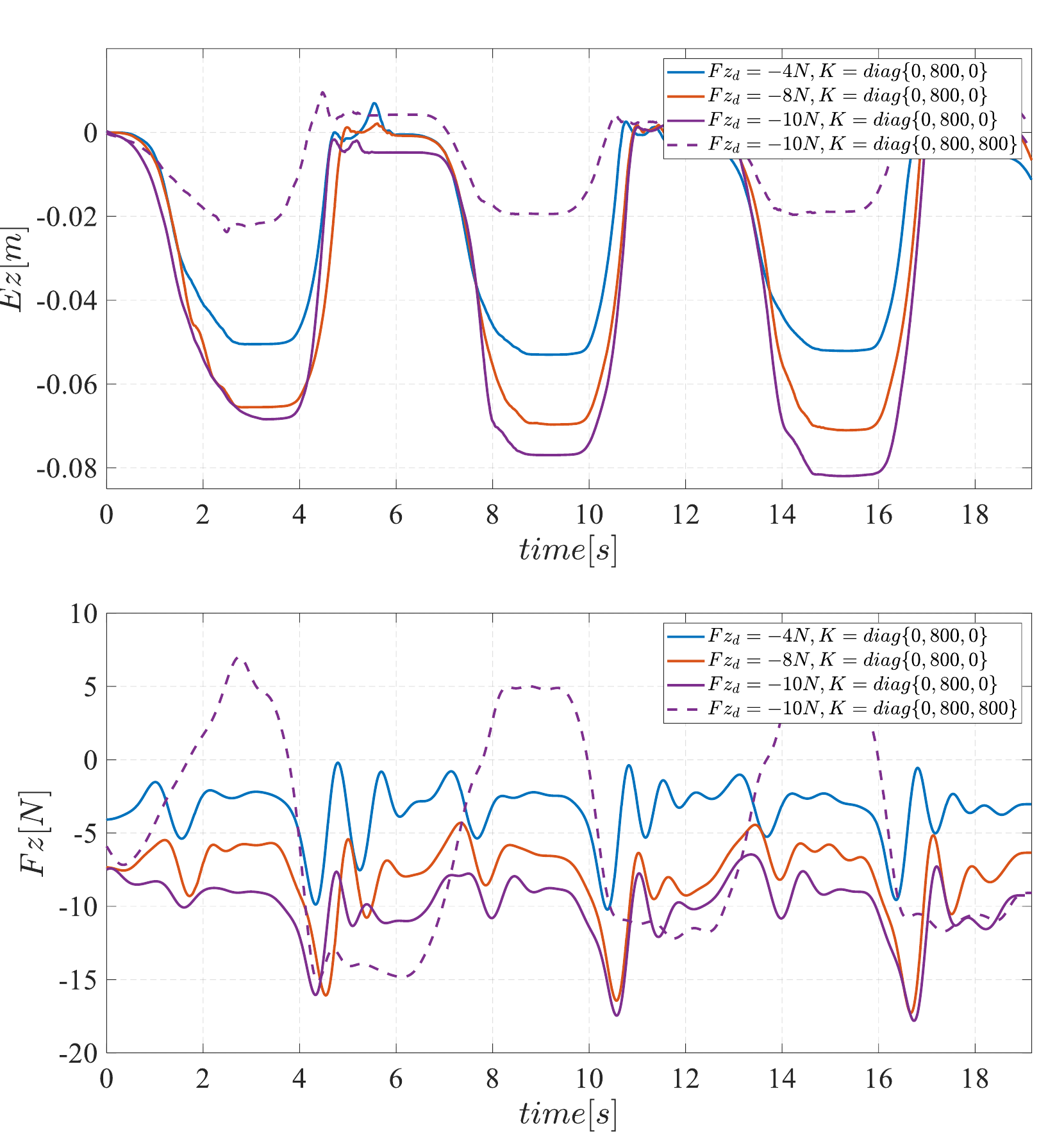}
    \caption{The top subplot illustrates that, when setting zero stiffness in $z$ direction, the tracking error $E_z$ becomes larger as the desired force $F_z$ increases. In our experiment when $F_z = -10N$, the saw rotated significantly due to the large $E_z$ and the stability was compromised. However, when setting the stiffness in $z$ direction to 800  $\nicefrac{N}{m}$, $E_z$ significantly decreased and instabilities were never observed. The bottom subplot shows the force tracking behaviors for the different settings.}
    \label{fig:comparison_z}
\end{figure}

\subsection{Robots' performance evaluation and validation}
To evaluate the potential of transferring cooperative dynamic manipulation skills from human demonstration videos to robots, we designed two sets of experiments: one to evaluate the contribution of stiffness in $z$ direction to the task stability, while the other, to evaluate the contribution of stiffness in $y$ direction (i.e., the main task direction) to the task performance and leader-follower relationship.

The robotic experimental setup is illustrated at the bottom of Fig.~\ref{fig:experiment_setup}. Two Franka Emika Panda robots were deployed in a center-symmetrical arrangement. Each side of the saw was grasped by the Pisa/IIT SoftHand, which was rigidly connected to each robot's end-effector. Both robots were controlled under a Cartesian impedance controller, based on the control framework presented in \cite{albu2007unified}. The sawing task setup was similar to the one described in human demonstrations. 

Our first observation from human demonstrations revealed that a high stiffness profile in $z$ direction can help to preserve the stability of the collaborative task. Hence, we performed a comparative experiment to validate this aspect. The experiment was conducted in different \textit{constant} stiffness (0 and 800 $\nicefrac{N}{m}$) and pushing force (-4, -8 and -10 $N$) settings in the $z$-direction, for both robots. 
The stiffness value in the sawing direction ($y$) was set to 800 $\nicefrac{N}{m}$ and kept fixed for both robots, to implement a trade off between the cutting force generation and robot compliance. This is because, a very compliant robot cannot produce the necessary reactive forces to overcome the friction between the saw blade and the surface, while a stiff profile will result in a non-harmonic collaborative action, since any possible trajectory mismatch can result in opposing forces between the two robots. Finally, since the sawing movement in $x$ direction was fully constrained, the stiffness value for both robots was set to zero to achieve a very complaint behaviour.

 

Fig.~\ref{fig:comparison_z} illustrates the results of the tracking error ($E$) and the reactive forces ($F$) in $z$ direction, while performing the sawing task under the different stiffness and forces conditions. As the desired force increased in $z$ direction, the tracking errors became larger and larger. 
In particular, when the desired force in $z$ direction was set to $-10N$, due to a large error in $z$ direction, the saw rotated significantly and the sawing task had to be interrupted due to this unstable behaviour. However, when the stiffness in $z$ direction was set to 800 $\nicefrac{N}{m}$, the desired force was still kept at $-10N$, and the the tracking performance showed the maximum tracking error of 0.02$m$ (purple dashed line in Fig.~\ref{fig:comparison_z}), in comparison to the with zero stiffness case, with the maximum tracking error of 0.08$m$ (purple solid line) in stable conditions. With reference to the force tracking behaviour in $z$ direction, as shown at the bottom of Fig.~\ref{fig:comparison_z}, it could be observed that the sawing force was much more stable with a high stiffness value. We can conclude that the force tracking behaviour with high stiffness in $z$ direction also outperformed the zero stiffness case.

\begin{figure}[b!]
    \vspace{-0.70cm}
    \centering
    \includegraphics[trim=0.0cm 1.0cm 1.0cm 1.0cm,clip,width=\linewidth]{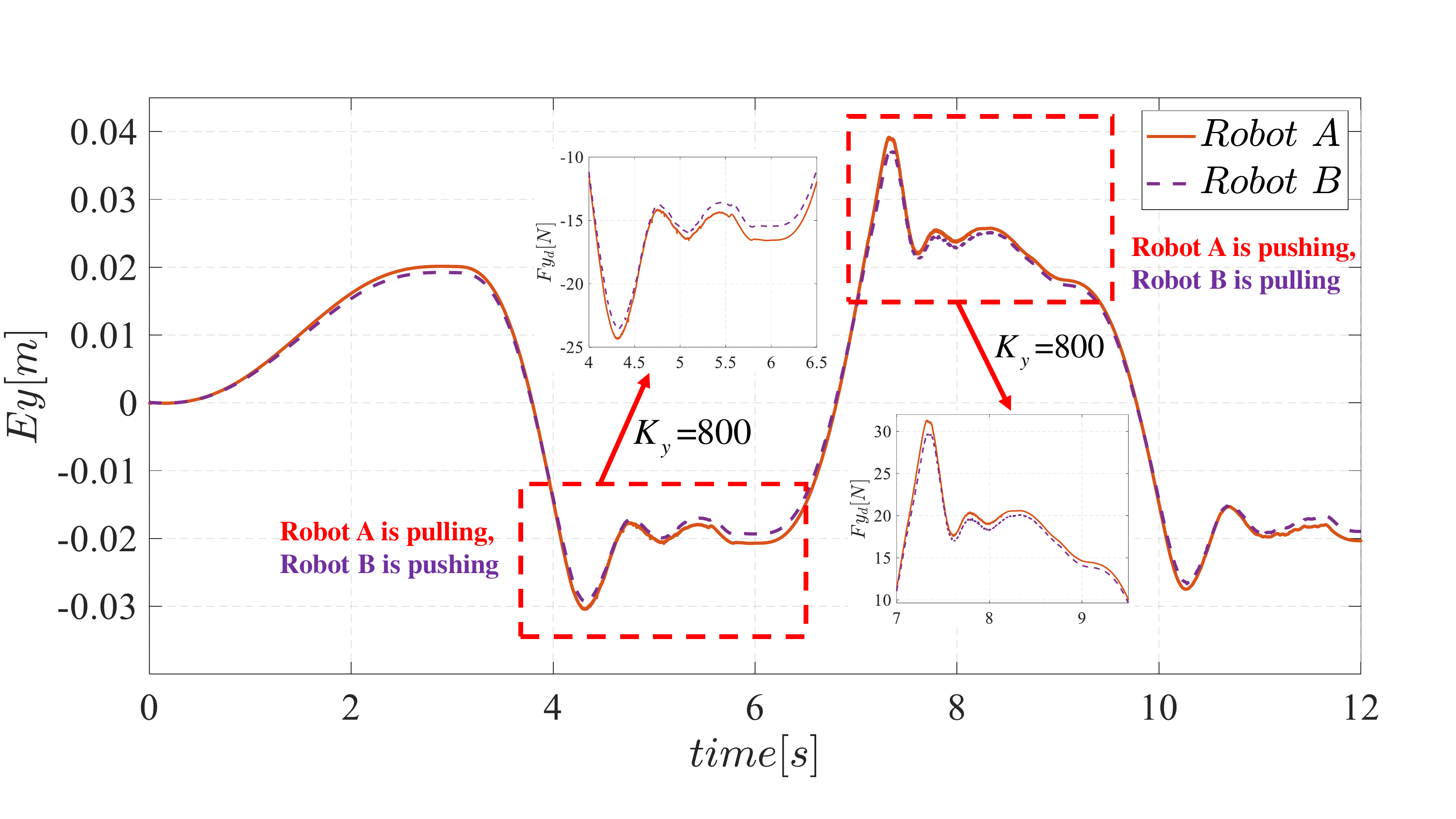}
    \caption{Similar constant stiffness values in $y$ direction for the two robots resulted in a similar tracking error ($E$) and the reaction forces, as a consequence. Hence, the leader-follower relation in this stiffness setting was far from being ideal. As observed in the plot, whenever \textit{Robot A} was pulling or pushing, its reactive force was always larger than that from \textit{Robot B}, which may cause discontinuity or even instability in the sawing motion when \textit{Robot A} was pushing while \textit{Robot B} was pulling since the pushing force was larger than the pulling force.}
    \label{fig:role_allocation_same_constant_stiffness}
\end{figure}

\begin{figure}[t!]
    \vspace{-0.1cm}
    \centering
    \includegraphics[trim=0.0cm 1.0cm 1.0cm 1.0cm,clip,width=\linewidth]{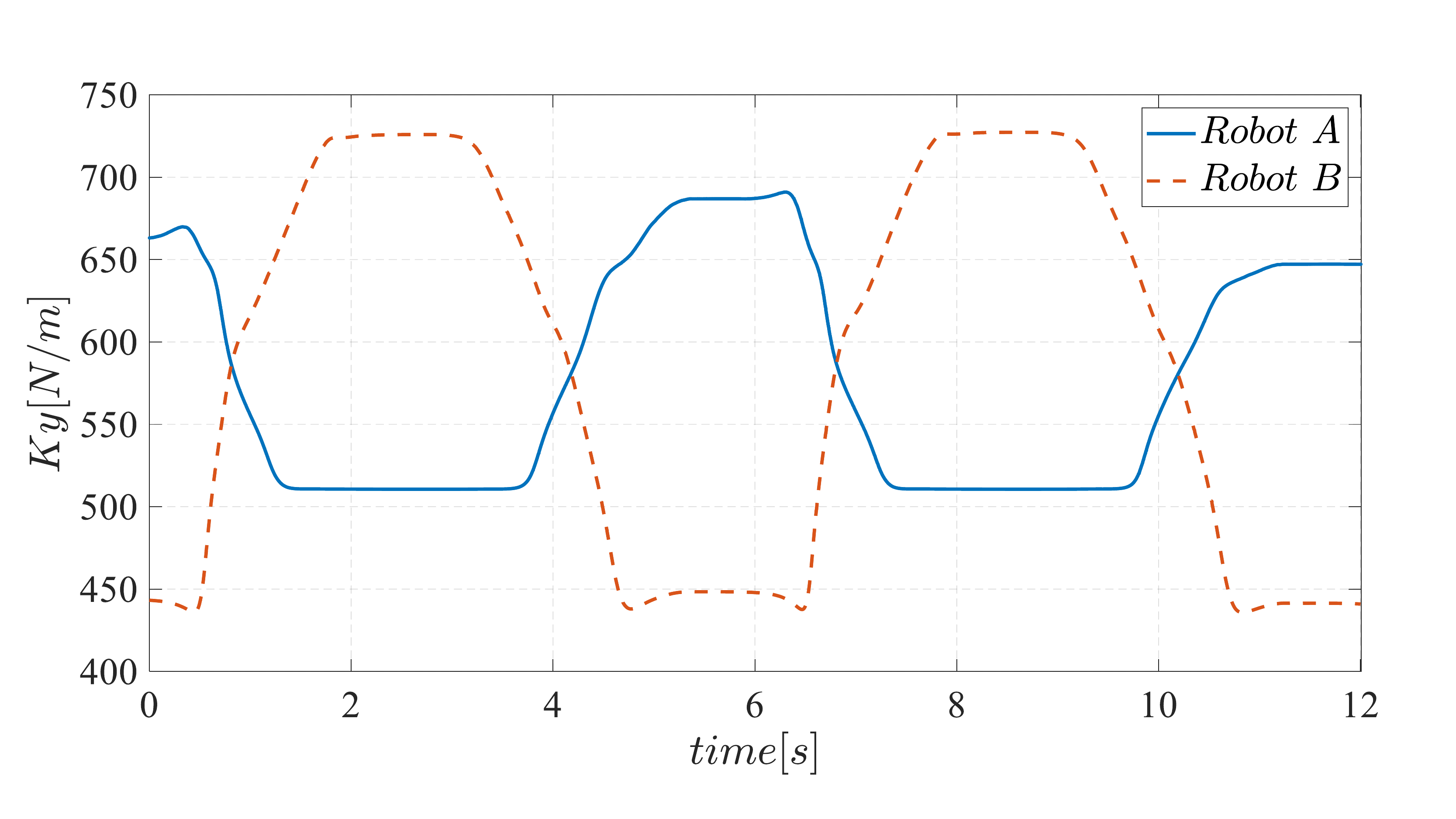}
    \caption{Learned coordinated stiffness profiles in $y$ direction from two subjects, taking $y$ and $z$ positions as input.}
    \label{fig:role_allocation_human_stiffness}
    \vspace{-0.5cm}
\end{figure}

\begin{figure}[t!]
\begin{subfigure}[b]{0.5\textwidth}
    \centering
    \includegraphics[trim=0.0cm 1.0cm 0.0cm 1.0cm,clip,width=\linewidth]{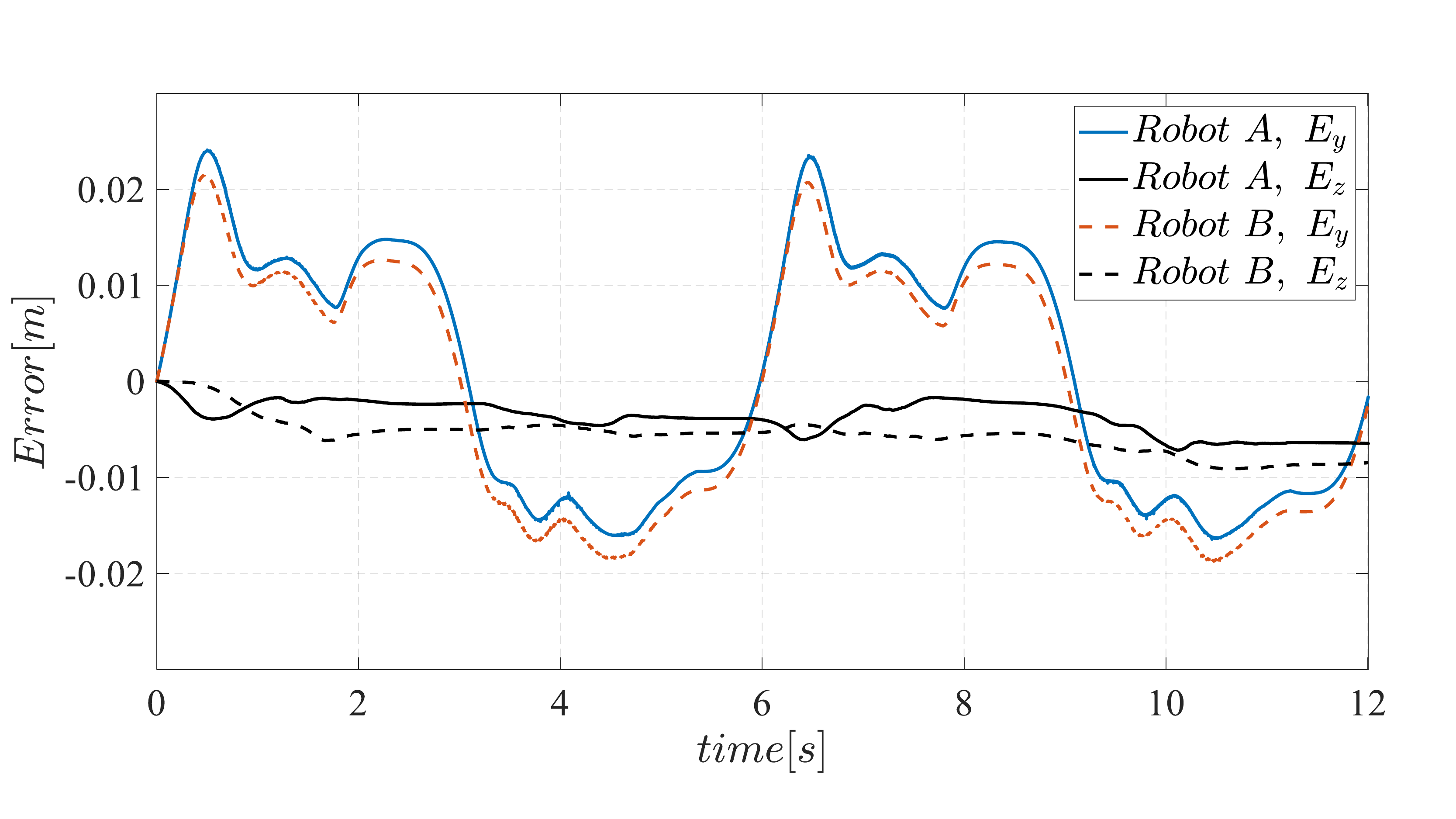}
    \caption{Tracking error.}
    \label{fig:role_allocation_human_error}
\end{subfigure}
\begin{subfigure}[b]{0.5\textwidth}
    \centering
    \includegraphics[trim=0.0cm 1.0cm 0.0cm 1.0cm,clip,width=\linewidth]{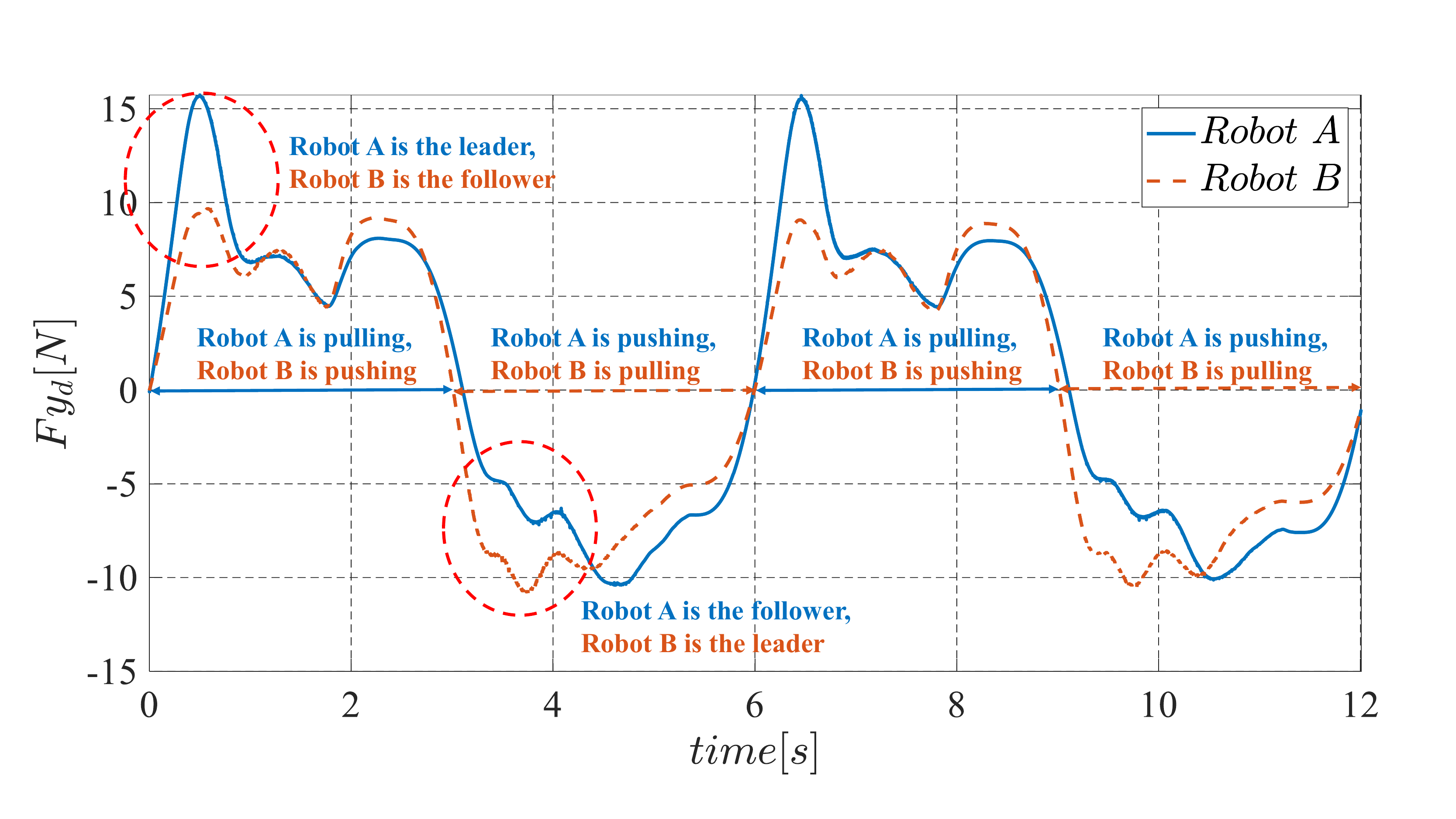}
    \caption{Reactive forces.}
    \label{fig:role_allocation_human_force}
\end{subfigure}
    \caption{When employing the coordinated stiffness profiles learned from humans in $y$ direction, similar tracking error in $y$ direction $E_y$ (on the top) for \textit{Robot A} and \textit{B} led to different reactive forces (at the bottom), resulting in an autonomous leader/follower role assignment and switching during the sawing task. The tracking error in $z$ direction $E_z$ by using the learnt CDS patterns (within 0.01m) was also smaller than that (around 0.02m) by using the same constant stiffness.}
    \vspace{-0.5cm}
\end{figure}

In the above set of experiments, constant stiffness values were set for both robots, leading to `non-ideal' collaborative behaviors for \textit{Robots A} and \textit{B}, especially in $y$ direction.
Fig.~\ref{fig:role_allocation_same_constant_stiffness} shows the tracking error ($E$) and the reactive forces ($F$) in $y$ direction for \textit{Robots A} (red line) and \textit{B} (dashed line) under the same constant stiffness condition (i.e., same condition as the dashed purple line in Fig.~\ref{fig:comparison_z}). The tracking error and the reaction forces applied by \textit{Robot A} were always slightly higher than \textit{Robot B} in certain moments. This difference was probably caused by the fact that, after we setup the robot experiment scenario, the goal position of the trajectory planner for each robot was decided by reading the current pose of robot's end-effector, in which there was minor uncertainty. The resulted larger reactive force from \textit{Robot A} whenever it is pulling or pushing therefore led the leader/follower role allocation issue. For instance, when \textit{Robot B} is pulling while \textit{Robot A} is pushing, it can be observed in Fig.~\ref{fig:role_allocation_same_constant_stiffness} that the reactive force from \textit{Robot A} was still larger than \textit{Robot B} which will cause discontinuity or even stability issues if the tracking error in $z$ direction is large enough in the sawing task.


To confront this issue and based on the observations of the leader-follower relationship in human experiments, 
we deployed the stiffness profile (encoded via CDS) in $y$ direction into the two robots' sawing process. This stiffness profile was learnt from human demonstrations, as illustrated in Fig.~\ref{fig:role_allocation_human_stiffness}. The resulting tracking errors and the reactive forces for the two robots are illustrated in Fig.~\ref{fig:role_allocation_human_error} and Fig.~\ref{fig:role_allocation_human_force} respectively. In the beginning of the experiment, \textit{Robot A} had a larger stiffness profile in comparison with \textit{Robot B}, and generated necessary pulling forces to cut the wood. \textit{Robot B} had more compliant behaviour on the other hand facilitated this action and did not contribute to creating larger pushing forces. Towards the end of this phase, the stiffness trends of the two robots changed, and hence their contribution to pulling and pushing actions, which determine the role of each robot in leading or following the collaborative phases. In fact, the transferring of the learnt CDS patterns in $y$ direction to the robots not only contributed to a better task performance in comparison to the constant stiffness case (judged by the tracking error in $z$ direction, the former is within 0.01m shown in Fig.~\ref{fig:role_allocation_human_error} while the latter is around 0.02m depicted in Fig.~\ref{fig:comparison_z}), but also achieved an effective role allocation and control sharing among the two.

\section{Discussions and Conclusions}
\label{sec:Conclusions}
In this work, the SPD stiffness matrices are pre-processed and converted into vectors by using Cholesky decomposition, and then the traditional Euclidean GMM/GMR is adopted to encode and reproduce human demonstrated stiffness profiles. It was shown in \cite{abu2018force,jaquier2021geometry} that Euclidean GMM/GMR with a Cholesky
decomposition could lead to inconsistent SPD matrices during
extrapolation. In the sawing experiment, it didn't exploit the extrapolation ability of GMM/GMR too much, also from Fig.~\ref{fig:cds_regressions}, it can be seen that the reproduced stiffness profile is consistent with the task requirements. However, if the proposed framework used in tasks that exploit extrapolation ability of GMM/GMR a lot, then the consistency problem has to be taken into account and the representation of stiffness matrices in GMM/GMR algorithms has to be based on Riemannian
geometry. 

In the sawing experiment, although the demonstrated stiffness is in 3D, only the value of stiffness ellipsoid along $y$ direction was transferred to robots. For $x$ direction, the saw is totally constrained by the wood, so it is reasonable to keep it as completely compliant. For $z$ direction, the stiffness in this direction keeps stability during sawing, due to the limitations of human arm itself, it can't always keep high stiffness in this direction. However, for robots, their behavior should not be limited because of this, to improve their performance in the sawing experiment, we set the stiffness is always high in $z$ direction. 

In summary, in this paper, we proposed a framework for learning cooperative dynamic manipulation  skills from human demonstration videos. Our motivation was to extend the concept of LfD to dynamic coordinates, benefiting from easily producible, large, and diverse  human demonstration datasets. We demonstrated the proposed method in a collaborative sawing task with leader-follower structure, considering environmental constraints and dynamic uncertainties. The experimental setup included two Panda robots, which replicated the leader-follower roles and the impedance profiles extracted from a two-persons sawing video. The aim was to tune robot impedance in such a highly coupled task not only to accomplish it but also to respond to possible trajectory mismatches with low interaction forces. Naturally, the hands' passive impedance behaviour also contributes to this, however, since robot compliance is higher, the interaction is majorly dominated by the arm impedance. These results revealed the potential of our approach in transferring of the key action principles from human demonstration videos to robots in dynamic industrial tasks.

\section{Acknowledgements}
\label{sec:Acknowledgements}
This work has been carried out in HRI$^{2}$ Lab, Istituto Italiano di Tecnologia, Genoa, Italy. It was supported in part by the ERC-StG Ergo-Lean [Grant Agreement No.850932], in part by the
National Natural Science Foundation of China under [Grant 91748208], and in part by the Department of Science and Technology of Shaanxi Province under [Grant 2017ZDL-G-3-1]. In this work, Y. Wu was also supported by the China Scholarship Council. \\


\bibliography{mybibfile}



\begin{wrapfigure}{l}{25mm} 
    \includegraphics[width=1in,height=1.25in,clip,keepaspectratio]{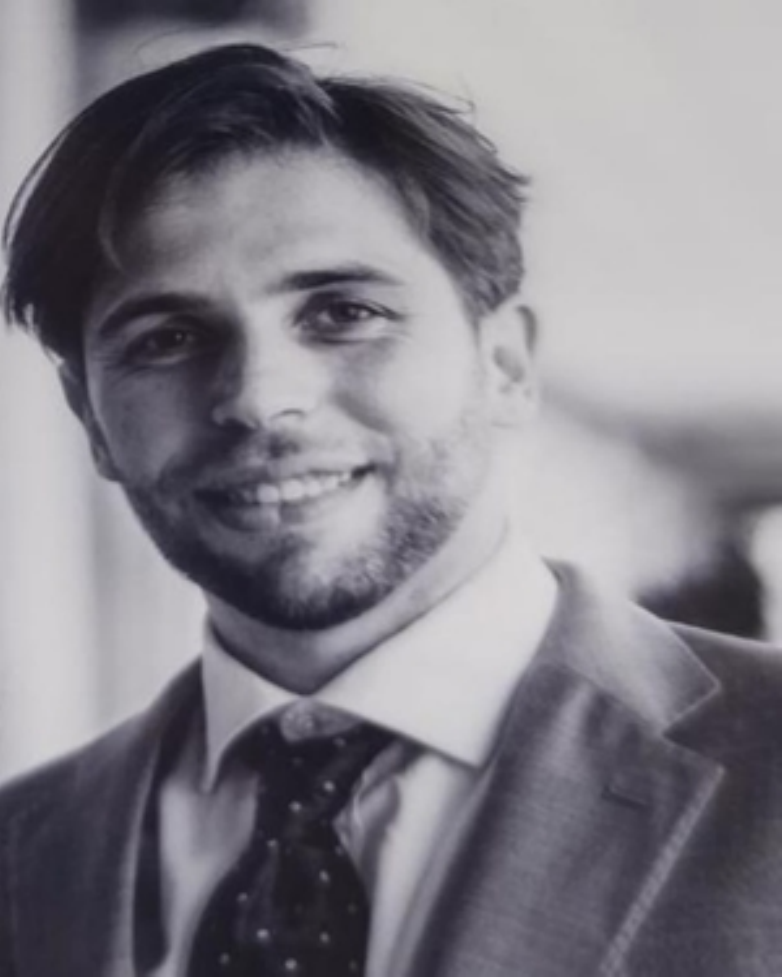}
 \end{wrapfigure}\par \textbf{Francesco Iodice} received the B.Eng. degree in computer engineering from Second University of Naples, Italy, in 2015. In October 2019, he obtained the Master of Science  in Artificial Intelligence and Robotics at Sapienza University of Rome.
Since 2020, he is pursuing a Ph.D. in Bioengineering at Politecnico di Milano, Neuroengineering and Medical Robotics Laboratory (NearLab), working in collaboration with Istituto Italiano di Tecnologia (IIT), under the supervision of Dr. Arash Ajoudani at Human-Robot Interfaces and Physical Interaction (HRI2) laboratory. His research concerns Computer Vision , Robotics and Machine Learning, and aims to improve human ergonomics in highly dynamic human-robot-environment interactions.\par
\vspace{1 cm}
\begin{wrapfigure}{l}{25mm} 
    \includegraphics[width=1in,height=1.25in,clip,keepaspectratio]{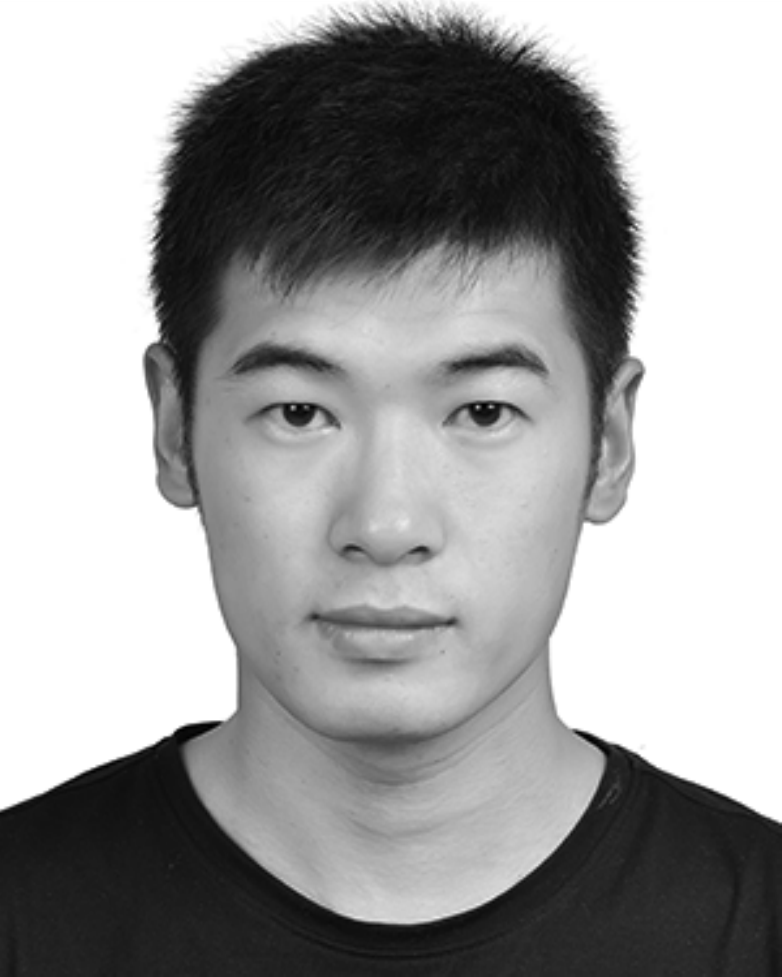}
 \end{wrapfigure}\par \textbf{Yuqiang Wu} received the B.Eng. degree in mechanical engineering from Xi’an Jiaotong University, Xi’an, China, in 2015. He is currently a PhD candidate in Xi’an Jiaotong University. From Oct 2018 to Oct 2020, He was a visiting PhD student at Human-Robot Interfaces and physical Interaction (HRI2) laboratory in Italian Institute of Technology (IIT). During that period, he started his research in mobile manipulation. His research interests include motion control of mobile manipulator, whole-body control, impedance/force control, human-like motor control, robot dynamics identification and computation. \par

\vspace{1 cm}
\begin{wrapfigure}{l}{25mm} 
    \includegraphics[width=1in,height=1.25in,clip,keepaspectratio]{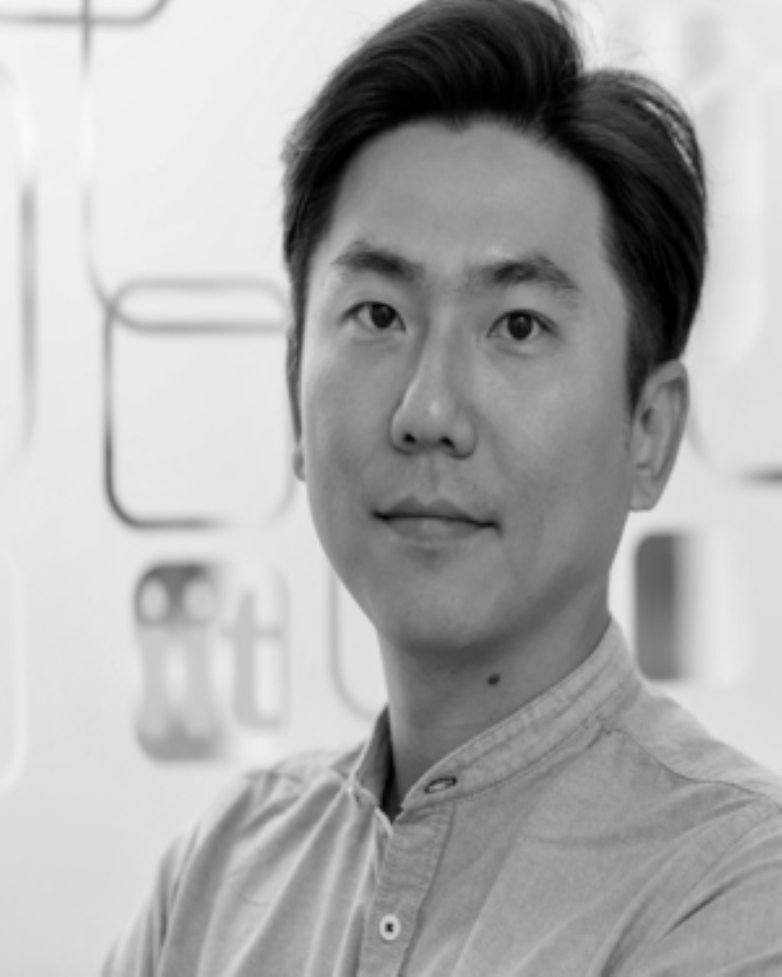}
 \end{wrapfigure}\par \textbf{Wansoo Kim} received the B.S. degree in mechanical engineering from Hanyang University ERICA campus, Korea in 2008 and PhD degree in mechanical engineering from Hanyang University, Korea in 2015 (Integrated MS/PhD program). He was a post-doctoral researcher at Human-Robot Interfaces and Physical Interaction Lab, Istituto Italiano di Tecnologia (IIT), Italy from 2016 to 2021. He is an assistant professor with Robotics Department, Hanyang University ERICA, Republic of Korea. He has developed several exoskeleton systems and collaborative robot (Cobot), and conducted research on the control of the powered exoskeleton robot and ergonomics collaboration control with Cobot through the physical human-robot interaction (pHRI) knowledge. He was involved in a Horizon-2020 project SOPHIA and an European Research Council project Ergo-lean. He has contributed to several projects in the field of exoskeleton robot in Korean projects (Development of Wearable Robot for Industrial Labor Support, etc.) He was the winner of the Solution Award 2019, the winner of the KUKA Innovation Award 2018, of the HYU best PhD paper award 2015, and of the ICCAS best presentation award 2014. His research interests are in Physical human-robot interaction (pHRI), human-robot collaboration, Shared Control, Ergonomics, Human modelling, Feedback devices, and powered exoskeleton robot.\par
 
\vspace{1 cm}
\begin{wrapfigure}{l}{25mm} 
    \includegraphics[width=1in,height=1.25in,clip,keepaspectratio]{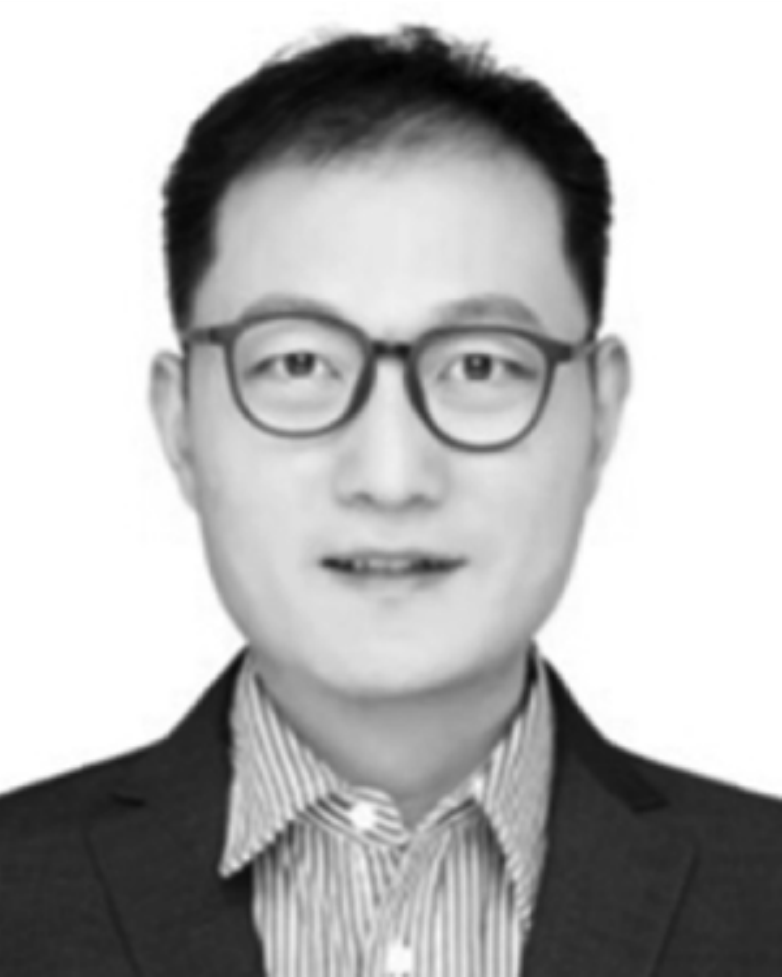}
 \end{wrapfigure}\par \textbf{Fei Zhao} (Member, IEEE) received the Ph.D. degree in mechanical engineering from Xi’an Jiaotong University, Xi’an, China, in 2013. He joined the School of Mechanical Engineering, Xi’an Jiaotong University, and the Shaanxi Key Laboratory of Intelligent Robots, in 2017. His research interests include robot kinematics and condition monitoring for intelligent manufacturing equipments.\par
\vspace{1 cm}
 \begin{wrapfigure}{l}{25mm} 
    \includegraphics[width=1in,height=1.25in,clip,keepaspectratio]{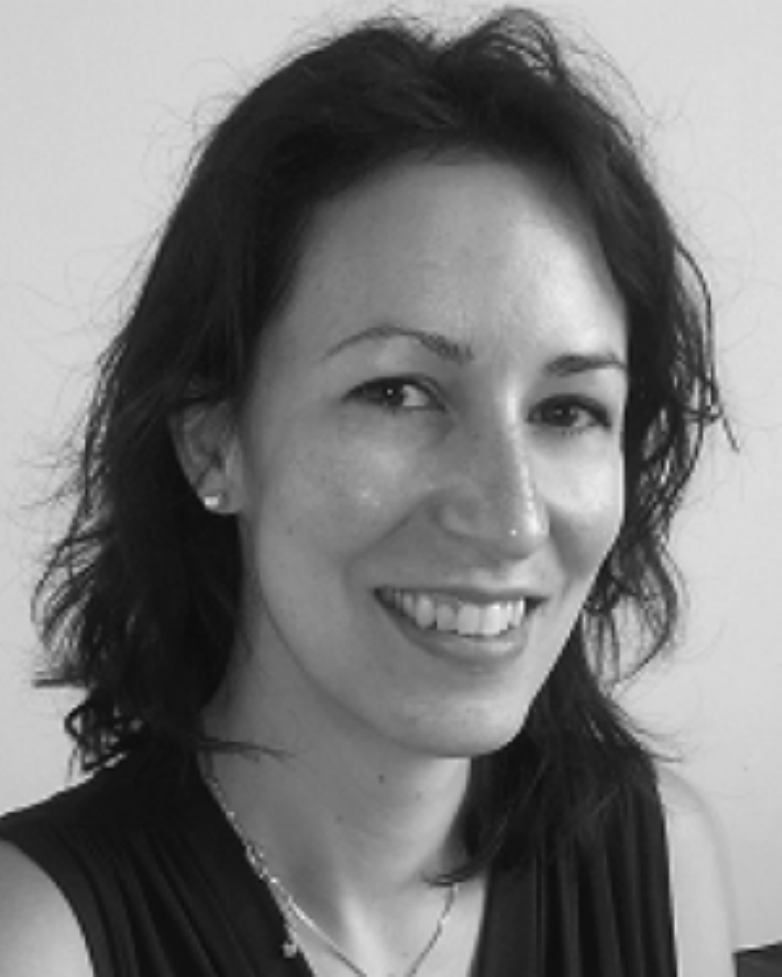}
 \end{wrapfigure}\par \textbf{Elena De Momi} (Senior Member, IEEE) received the M.Sc. and Ph.D. degrees in biomedical engineering from the Politecnico di Milano, Milan, Italy, in 2002 and 2006, respectively.,She is an Associate Professor with the Department of Electronics, Information and Bioengineering, Politecnico di Milano. She is a Co-Founder of the Neuroengineering and Medical Robotics Laboratory, in 2008, where she is responsible for the Medical Robotics Section., Dr. De Momi is currently an Associate Editor of the Journal of Medical Robotics Research, the International Journal of Advanced Robotic Systems, Frontiers in Robotics, and AI and Medical {\&} Biological Engineering {\&} Computing. From 2016, she has been an Associate Editor of IEEE International Conference on Robotics and Automation (ICRA), International Conference on Intelligent Robots and Systems (IROS), and BioRob. She is currently a Publication Co-Chair of ICRA 2019. She is responsible for the lab course in Medical Robotics and of the course on Clinical Technology Assessment of the M.Sc. degree in Biom. Eng. at Politecnico di Milano, and she serves on the board committee of the Ph.D. course in bioengineering.\par
\vspace{2 cm}
\begin{wrapfigure}{l}{25mm} 
    \includegraphics[width=1in,height=1.25in,clip,keepaspectratio]{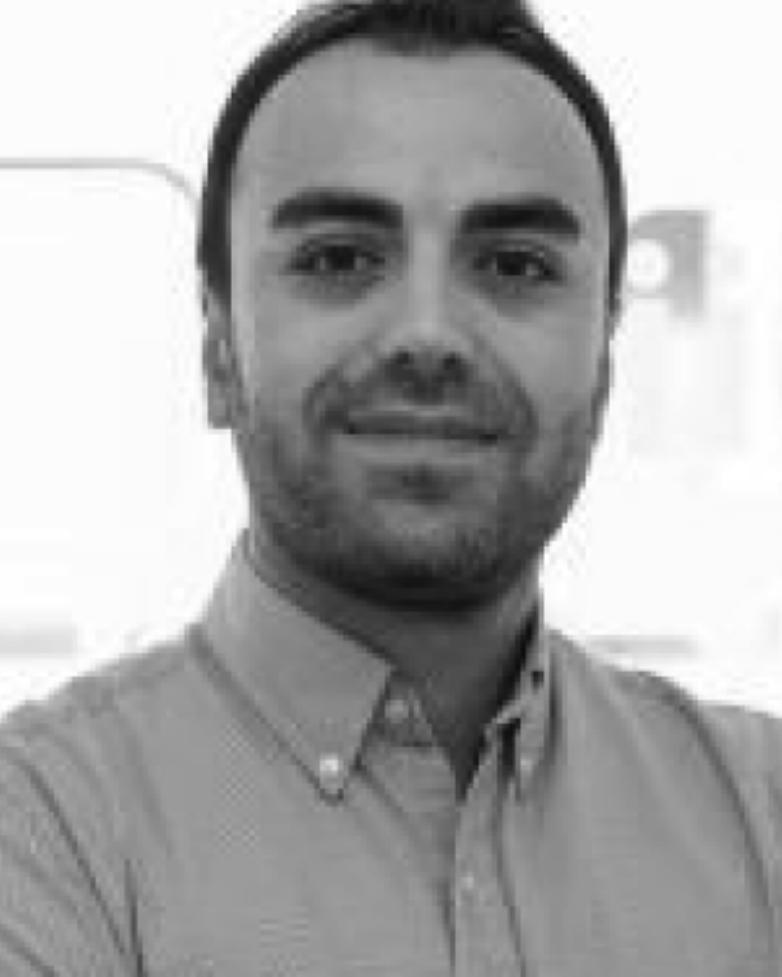}
 \end{wrapfigure}\par \textbf{Arash Ajoudani} (Member, IEEE) received the Ph.D. degree in robotics and automation from the University of Pisa and IIT, in July 2014. He is currently a tenured Senior Scientist with the Italian Institute of Technology (IIT), where he leads the Human-Robot Interfaces and physical Interaction (HRI2) Laboratory. He is the author of the book Transferring Human Impedance Regulation Skills to Robots [Springer Tracts in Advanced Robotics (STAR)], and several publications in journals, international conferences, and book chapters. His main research interests include physical human–robot interaction and cooperation, robotic manipulation, robust and adaptive control, assistive robotics, and tele-robotics. He has contributed to several successful European projects (H2020 and FP7), such as WALKMAN, WearHap, SOMA, and SoftPro. He was a recipient of the European Research Council (ERC) starting grant 2019. His Ph.D. Thesis was a Finalist for the Georges Giralt Ph.D. Award 2015–Best European Ph.D. Thesis in robotics. He was a Winner of the Amazon Research Awards, in 2019, the Solution Award, in 2019 (Premio Innovazione Robotica at MECSPE2019), the KUKA Innovation Award, in 2018, and the Werob Best Poster Award, in 2018; a Finalist for the Best Manipulation Paper Award at ICRA, in 2012, the Best Conference Paper Award at Humanoids, in 2018, the Best Interactive Paper Award at Humanoids, in 2016, and the Best Oral Presentation Award at Automatica (SIDRA), in 2014; and a Winner of the Best Student Paper Award and a Finalist for the Best Conference Paper Award at ROBIO, in 2013. He is the coordinator of the Horizon-2020 Project SOPHIA with a consortium of 12 partners from the leading European research and industrial organizations. He has been serving as the Executive Manager of the IEEE-RAS Young Reviewers’ Program (YRP), and the Chair and a Representative of the IEEE-RAS Young Professionals Committee. He has been serving as a member for the Scientific Advisory Committee and an Associate Editor for several international journals and conferences, such as IEEE RAL, Biorob, and ICORR.\par

\end{document}